*Article*

# CORE-ReID: Comprehensive Optimization and Refinement through Ensemble fusion in Domain Adaptation for person re-identification

**Trinh Quoc Nguyen [1,2,*], Oky Dicky Ardiansyah Prima [1,*] and Katsuyoshi Hotta [2]**

[1] Graduate School of Software and Information Science, Iwate Prefectural University; g236v201@s.iwate-pu.ac.jp (T.Q.N.); prima@iwate-pu.ac.jp (O.D.A.P.)
[2] CyberCore Co., Ltd; hotta@cybercore.co.jp (K.H.)
* Correspondence: g236v201@s.iwate-pu.ac.jp (T.Q.N.)

**Abstract:**

This study introduces a novel framework, "Comprehensive Optimization and Refinement through Ensemble Fusion in Domain Adaptation for Person Re-identification (CORE-ReID)", to address an Unsupervised Domain Adaptation (UDA) for Person Re-identification (ReID). The framework utilizes CycleGAN to generate diverse data that harmonizes differences in image characteristics from different camera sources in the pre-training stage. In the fine-tuning stage, based on a pair of teacher-student networks, the framework integrates multi-view features for multi-level clustering to derive diverse pseudo labels. A learnable Ensemble Fusion component that focuses on fine-grained local information within global features is introduced to enhance learning comprehensiveness and avoid ambiguity associated with multiple pseudo-labels. Experimental results on three common UDAs in Person ReID demonstrate significant performance gains over state-of-the-art approaches. Additional enhancements, such as Efficient Channel Attention Block and Bidirectional Mean Feature Normalization mitigate deviation effects and adaptive fusion of global and local features using the ResNet-based model, further strengthening the framework. The proposed framework ensures clarity in fusion features, avoids ambiguity, and achieves high accuracy in terms of Mean Average Precision, Top-1, Top-5, and Top-10, positioning it as an advanced and effective solution for the UDA in Person ReID. Our codes and models are available at https://github.com/TrinhQuocNguyen/CORE-ReID.

## 1. Introduction

In the context of Person Re-identification (ReID) [1], where the objective is to match images of individuals (e.g., pedestrian, suspect, etc.) across non-overlapping camera views, the importance of efficient and accurate identification holds significant implications for applications in smart cities and large-scale surveillance systems [2-4]. Recent advancements in deep learning techniques have shown promising improvements in ReID performance [5,6]. However, these techniques often require a substantial amount of labeled data for effective training, which limits their applicability in real-world settings. The reliance on labeled data for training poses constraints, particularly in scenarios where manual labeling is resource-intensive and expensive.

The inherent limitations of supervised strategies stem from the need for manually labeled cross-view training data, a resource-intensive process that incurs significant expenses [7,8]. In Person ReID, these limitations become particularly marked due to two primary reasons: (1) the reliability of manual labeling diminishes when dealing with a large number of images across multiple camera views, and (2) the exorbitant cost in terms of both time and money poses a formidable barrier to labeling the vast amount of data spanning disjoint camera views [9]. Consequently, in practical scenarios, the applicability of supervised methods is limited, especially when confronted with a substantial amount of unlabeled data in a new context.

A viable solution is considered by addressing the challenge of adapting a model trained on a labeled source domain to an unlabeled target domain, known as the Unsupervised Domain Adaptive (UDA) in Person ReID. However, it remains a formidable task due to the existing data distribution gap and the presence of non-overlapping identities between the source and target domains. Notably, methods of prevalent the UDA Person ReID [10-13], often referred to as "fine-tuning", first pre-train the model on the labeled source domain, then perform clustering algorithms and similarity measurements to generate pseudo labels on the unlabeled target domain to further refine the model. Despite their effectiveness in improving performance in the target domain, these methods tend to overlook the influence of camera variations in the source domain, which significantly affect the performance of the pre-trained model before entering the fine-tuning stage. Our approach is consistent with the concept of the "fine-tuning" approach, but we emphasize increasing camera awareness in the initial stage when training the model using the labeled data. Our motivation stems primarily from the necessity for a large amount of data in deep learning-based Person ReID. Annotating large-scale datasets to develop reliable features that can handle camera variations is beneficial. However, it is also prohibitively expensive.

This study introduces a Comprehensive Optimization and Refinement through Ensemble Fusion in Domain Adaptation for Person Re-identification (CORE-ReID) framework to refine the model on the target domain dataset during the fine-tuning stage. While Self-Similarity Grouping (SSG) [14] and Learning Feature Fusion ($LF^2$) [15] explore the use of both local and global features of the UDA in Person ReID, they have certain challenges. First, SSG uses a single network for feature extraction in clustering, which is susceptible to the generation of numerous noisy pseudo-labels. In addition, it performs clustering based on global and local features independently, resulting in unlabeled samples acquiring multiple different pseudo-labels, leading to ambiguity in identity classification during training. Second, $LF^2$ adopts a similar approach to the channel attention module of the Convolutional Block Attention Module (CBAM) [16] in its fusion module. However, the simplicity of CBAM's design may not be optimal. Moreover, $LF^2$ does not optimize the final features using horizontally flipped images, which may result in suboptimal attention maps for distinguishing identity-related features from background features. In contrast, the CORE-ReID addresses these limitations by incorporating horizontally flipped images and employing an Efficient Channel Attention Block (ECAB). The advantage of ECAB is that it enhances feature representation through attention mechanisms that emphasize important and deterministic features when performing re-identification. Our model utilizes a channel attention map by exploiting inter-channel relationships within features, which will serve as a feature detector. In addition, Bidirectional Mean Feature Normalization (BMFN) is used to fuse features from both original and flipped images. These enhancements make CORE-ReID a promising approach to bridging the gap between supervised and unsupervised methods in Person ReID and provide valuable insights into UDA for this domain, potentially paving the way for future advancements.

Experimental results conducted on three widely used UDA Person ReID datasets demonstrate that our method outperforms state-of-the-art approaches in terms of performance. To summarize, our study makes the following major contributions:

- **Novel Dynamic Fine-Tuning Approach with Camera-Aware Style Transfer:** We introduce a pioneering fine-tuning strategy that employs a camera-aware style transfer model for Re-ID data augmentation. This novel approach not only addresses disparities in images captured by different cameras but also mitigates the impact of Convolutional Neural Network (CNN) overfitting on the source domain.
- **Innovative Efficient Channel Attention Block (ECAB):** We develop a groundbreaking ECAB that leverages inter-channel relationships of features to guide the model's attention to meaningful structures within the input image. This innovation enhances feature extraction and focuses the model on critical identity-related features.

- **CORE Framework with Ensemble Fusion of Global and Local Features:** We establish the CORE (Comprehensive Optimization and Refinement through Ensemble fusion) framework, which utilizes a novel pair of teacher-student networks to perform an adaptive fusion of global and local (top and bottom) features for multi-level clustering with the objective of generating diverse pseudo-labels. By proposing the Bidirectional Mean Feature Normalization (BMFN), the model can increase its discriminability at the feature level and address key limitations in existing methods.

## 2. Related Work

This chapter visits the related research work, including the field of the Unsupervised Domain Adaptation (UDA) for Person ReID and the knowledge transfer through methods like knowledge distillation, highlighting approaches that aim to transfer expertise from well-trained models to enhance learning and adaptation in challenging domain scenarios. We can categorize the UDA methods into three main groups: style-transferred source-domain images, clustering-based approaches, and feature alignment methods. Each category presents unique strategies and challenges in adapting models to different domains, showcasing innovative techniques such as style transfer for domain-invariant features, iterative clustering for refined representations, and attribute alignment for knowledge transfer. Despite their successes, these methods face obstacles like image quality dependencies, noise in pseudo-labels, and domain shift adaptability.

### *2.1. Unsupervised domain adaptation for Person ReID*

The UDA has attracted significant interest due to its ability to reduce the need for costly manual annotation. This method can effectively leverage labeled data from a source domain to improve performance on a target domain without requiring annotations specific to the target domain. Generally, the UDA falls into three main categories: style-transferred source-domain images, clustering-based, and feature alignment methods.

**Style-transferred source-domain images:** This category focuses on learning domain-invariant features using style-transferred source-domain images. The main idea is to transfer low and mid-level target domain features, such as background, illumination, resolution, and clothing to the images in the source domain. Techniques such as SPGAN [17], PTGAN [18], and PDA-Net [19] operate by transferring source-domain images to mimic the visual style of the target domain while preserving the underlying person identities. These style-transferred images, along with their corresponding identity labels, are then used to fine-tune the model for improved performance on the target domain. Another notable approach within this category is Hetero-Homogeneous Learning (HHL) [20], which focuses on learning camera-invariant features through the use of camera style-transferred images. By mitigating the influence of camera-specific variations, HHL aims to increase the model's ability to handle domain shifts. However, despite their effectiveness in matching visual styles across domains, these methods have limitations. The retrieval performance of these methods is highly dependent on the quality of the generated images. Additionally, these approaches often overlook the intricate relationships between different samples within the target domain, limiting their ability to capture the complex dynamics present in real-world scenarios.

**Clustering-based methods:** The second category, clustering-based approaches, continues to maintain state-of-the-art performance in the field. In particular, Fan et al. [12] introduced a method that alternately assigns labels to unlabeled training samples and optimizes the network using the generated targets. This iterative process facilitates effective domain adaptation by iteratively refining the model's representations to match with the target domain. Building on this foundation, Lin et al. [21] proposed a bottom-up clustering framework supplemented by a repelled loss mechanism. This approach aims to improve the discriminative power of the learned representations while mitigating the effects of intra-cluster variation. Similarly, SSG [14] and LF$^2$ [15] contributed to this category by introducing a technique that assigns pseudo-labels to both global and local features. Ge et al. [22] introduced the Mutual Mean-Teaching (MMT) method, which uses off-line refined hard pseudo-labels and on-line refined soft pseudo-labels in an alternative training approach to learn enhanced features from the target domain. This innovative method enhances the model's ability to adapt to domain shifts by iteratively refining pseudo-labels and feature representations during training. In addition, Zheng et al. [23] established the Uncertainty-Guided Noise-Resilient Network (UNRN), which explores the credibility of predicted pseudo-labels of target domain samples. By considering uncertainty estimates in the training process, UNRN increases the model's performance to noisy annotations and improves its performance in domain adaptation scenarios. By using information from both levels of abstraction, these methods achieve improved performance in capturing fine-grained distinctions within the target domain. However, despite their success, clustering-based methods face challenges related to the noise inherent in the hard pseudo-labels generated by clustering algorithms. This noise can significantly hinder the training of neural networks and is often not addressed by existing methods.

**Feature alignment:** The third category of domain adaptation methods aims to align common attributes in both source and target domains to facilitate knowledge transfer. These attributes may include clothing items and other soft-biometric characteristics that are common in both domains. By aligning mid-level features associated with these attributes, these methods enable the learning of higher-level semantic features in the target domain. For instance, works such as TJ-AIDL [24] consider a fixed set of attributes for alignment. To enhance generalization capabilities, Lin et al. [25] propose the Multi-task Mid-level Feature Alignment (MMFA) technique. MMFA enables the method to learn attributes from both domains and align them for improved generalization on the target domain. Furthermore, UCDA [26] and CASCL [27] aim to align attributes by considering images from different cameras within the target dataset. Wang et al. [24] proposed a model capable of simultaneously learning an attribute-semantic and identity-discriminative feature representation space that is transferable to the target domain. This method contributes to advancing domain adaptation by effectively aligning attribute-level features and improving the transferability of learned representations between domains. However, challenges arise due to differences in pedestrian classes between the two domains, making it difficult for the model to learn a common feature representation space.

### *2.2. Knowledge transfer*

Knowledge distillation, the process of transferring knowledge from a well-trained neural network (often referred to as the teacher model) to another model or network (referred to as the student model), has received significant attention in recent years [28-30]. The fundamental concept involves creating consistent training supervisions for labeled and unlabeled data through the predictions of different models. For instance, the mean-teacher model introduced in [31] innovatively averages model weights across various training iterations to generate supervisions for unlabeled samples. In contrast, Deep Mutual Learning, proposed by Zhang et al. [32], diverges from the traditional teacher-student paradigm by employing a pool of student models. These student models are trained collaboratively and provide supervision to each other, thus promoting mutual learning and exploration of different representations. Ge et al. proposed MMT [22], which adopts an alternative training approach utilizing both off-line refined hard pseudo-labels and on-line refined soft pseudo-labels. MEB-Net [33] utilizes three networks (six models) to perform mutual mean teacher training to generate the pseudo labels. However, despite their effectiveness, these methods face challenges. They rely heavily on pseudo-labels generated by the teacher network, which may be inaccurate or noisy, leading to suboptimal performance. Additionally, they may struggle to adapt effectively to significant domain shifts, especially when domains exhibit significant differences in lighting conditions, camera viewpoints, or background clutter, resulting in degraded performance.

## 3. Materials and Methods

We adopt the clustering-based method by separating the process into two stages: pre-training the model on the source domain in a fully supervised manner and fine-tuning the model on the target domain using an unsupervised learning approach (Figure 1). Our algorithm leverages a pair of teacher-student networks [34]. After training the model using a customized source domain dataset, the parameters of this pre-trained model will be copied to student and teacher networks as an initialized step to prepare for the next stage. At the fine-tuning stage, we train the student model and then optimize the teacher model using the Nesterov momentum. To reduce the computation cost, only the teacher model will be used for inference.

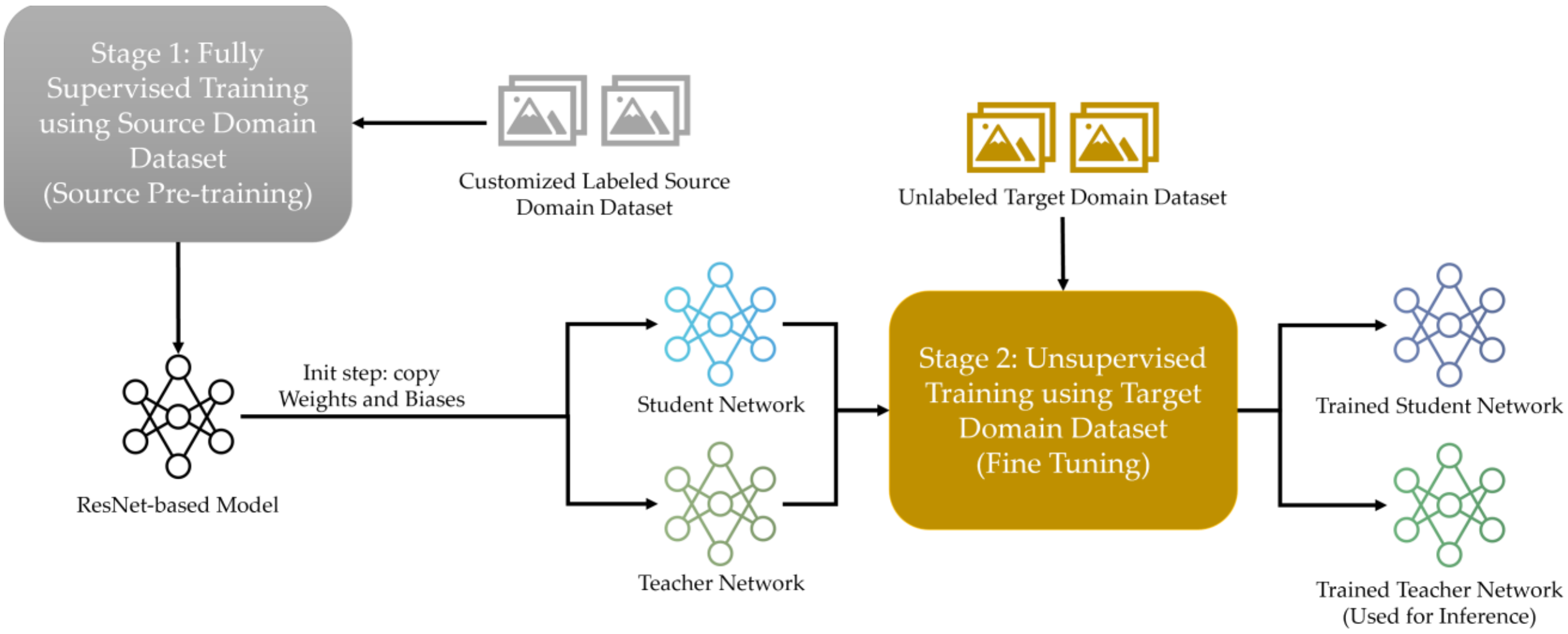

**Figure 1**. The model proposed in this study. First, the model is trained on a customized source domain dataset, and subsequently, the parameters of this pre-trained model are transferred to both the student and teacher networks as an initialization step for the next stage. During fine-tuning, we train the student model and then update the teacher model using momentum updates. To optimize computational resources, only the teacher model is used for inference purposes.

### *3.1. Camera-aware Image-to-Image translation on source domain dataset*

For any two unordered image collections $X$ and $Y$, comprised of training samples $\{x_i\}_{i=1}^{N}$ where $x_i \in X$ and $\{y_j\}_{j=1}^{M}$ where $y_j \in Y$, their respective data distributions are denoted as $x \sim p_{data}(x)$ and $y \sim p_{data}(y)$. When learning to translate images from a source domain $X$ to a target domain $Y$, the objective of CycleGAN [35] is to acquire a mapping $G: X \rightarrow Y$ that renders the distribution of images by $G$ indistinguishable from the distribution $Y$, which is achieved through an adversarial loss. Due to the inherent under-constraint of this mapping, CycleGAN [35] introduces an inverse mapping $F: Y \rightarrow X$, incorporating a cycle consistency loss to enforce $F(G(X)) \approx X$ and vice versa. CycleGAN further employs two adversarial discriminators, $D_X$ and $D_Y$, where $D_X$ discerns between images $\{x\}$ and translated images $\{F(y)\}$, and $D_Y$ distinguishes $\{y\}$ from $\{G(x)\}$. By leveraging the GAN framework, CycleGAN jointly trains generative and discriminative models. The comprehensive CycleGAN loss function is expressed as:

$$\mathcal{L}_{CycleGAN}(G, F, D_X, D_Y) = \mathcal{L}_{GAN}(G, D_Y, X, Y) + \mathcal{L}_{GAN}(F, D_X, Y, X) + \lambda \mathcal{L}_{cyc}(G, F), \quad (1)$$

where $\mathcal{L}_{GAN}(G, D_Y, X, Y)$ and $\mathcal{L}_{GAN}(F, D_X, Y, X)$ are two loss functions, corresponding to the mapping functions $G$ and $F$, as well as the discriminators $D_Y$ and $D_X$. Additionally, $\mathcal{L}_{cyc}(G, F)$ represents the *cycle consistency loss*, compelling the reconstructions $F(G(X)) \approx X$ and $G(F(Y)) \approx Y$ for each image after a cycle mapping. The parameter $\lambda$ serves to penalize the significance attributed to $\mathcal{L}_{GAN}$ compared to $\mathcal{L}_{cyc}$. This ensures a balanced consideration of adversarial and cycle consistency aspects. The method aims to solve:

$$G^*, F^* = \arg \min_{G,F} \max_{D_X, D_Y} \mathcal{L}(G, F, D_X, D_Y). \quad (2)$$

Further insights into the specifics of the CycleGAN framework can be found in [35].

Inspired by the CamStyle [36], we incorporate CycleGAN to generate new training samples, treating styles across different cameras as different domains. This approach involves learning image-image translation models using CycleGAN for images from different camera views $C$ in the Person ReID dataset. To ensure color consistency between input and output during style transfer, similar to the painting → photo application, we add an identity mapping loss proposed in [37] to the CycleGAN loss function (Eq. 1). This additional loss term compels the generator to approximate an identity mapping when real images from the target domain are used as input, expressed as:

$$\mathcal{L}_{identity}(G,F) = \mathbb{E}_{y\sim p_{data}(y)}[||G(y) - y||_1] + \mathbb{E}_{x\sim p_{data}(x)}[||F(x) - x||_1]. \quad (3)$$

The absence of $\mathcal{L}_{identity}$ would grant the generator G and F the freedom to alter the tint of input images unnecessarily. As the results, the total loss used for training is:

$$\mathcal{L}_{total}(G,F,D_X,D_Y) = \mathcal{L}_{CycleGAN}(G,F,D_X,D_Y) + \mathcal{L}_{identity}(G,F). \quad (4)$$

Our approach differs from CamStyle, which addresses Person ReID by utilizing and generating more data on the training dataset and evaluating the model within the same dataset (e.g., Market-1501, CUHK03). We aim to train on a source domain $S$ and evaluate the algorithm during the fine-tuning stage on a different target domain $T$. This approach allows us to leverage the entirety of the data in $S$ by incorporating test data into the training set, similar to that of DGNet++ [38].

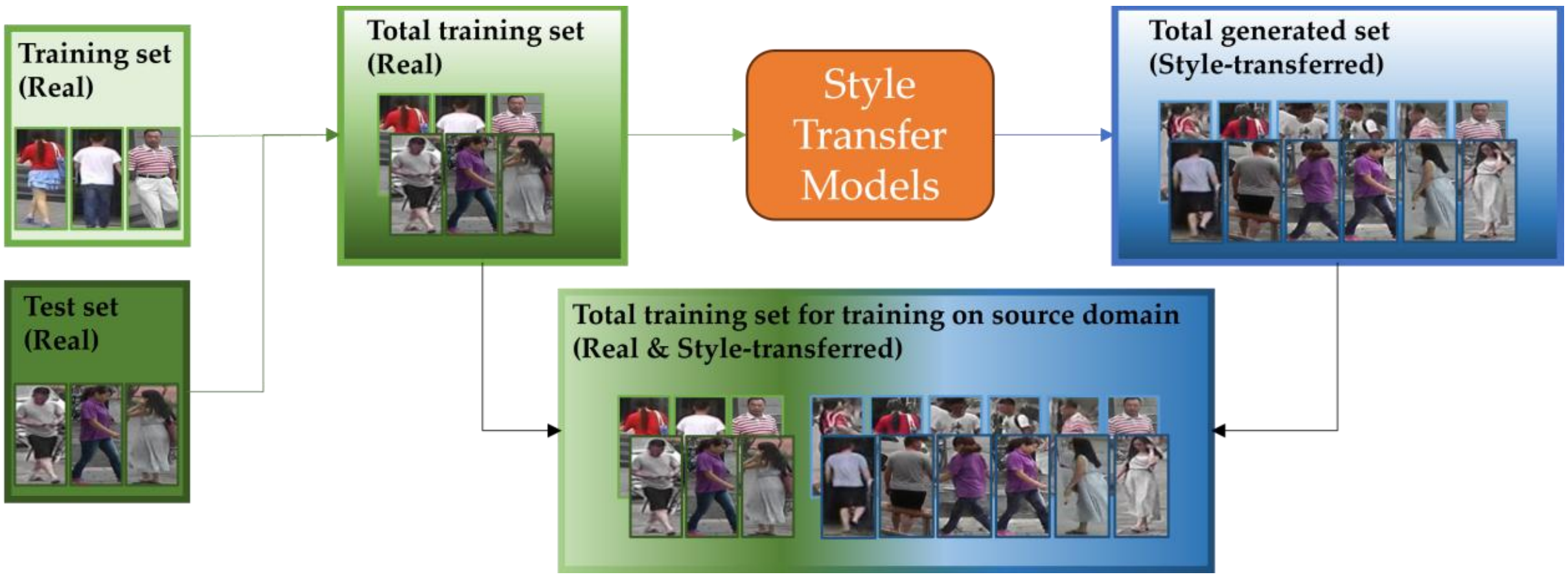

**Figure 2**. Our pipeline of creating the full training set for the source domain. Initially, we combine both the training set (**green** boxes) and the test set (**dark green** boxes) within the source dataset to form the total training set consisting of real images. This combined set is then used to train the camera-aware style transfer model. For each real image, the trained transfer model is applied to generate images (**blue** boxes for the training set and **dark blue** boxes for the test set) that align with the stylistic characteristics of the target cameras. Subsequently, the real images (**green** and **dark green** boxes) and the style-transferred images (**blue** and **dark blue** boxes) are merged to produce the final training set within the source domain.

For a source domain dataset containing images from $C$ different cameras, the number of generative models used to produce data both $X \rightarrow Y$ and $Y \rightarrow X$ is $C(C-1)$. Consequently, the final training set comprises a blend of the original real images and the style-transferred images from both the training and test sets within the source domain dataset (Figure 2). These style-transferred images seamlessly adopt the labels from the original real images.

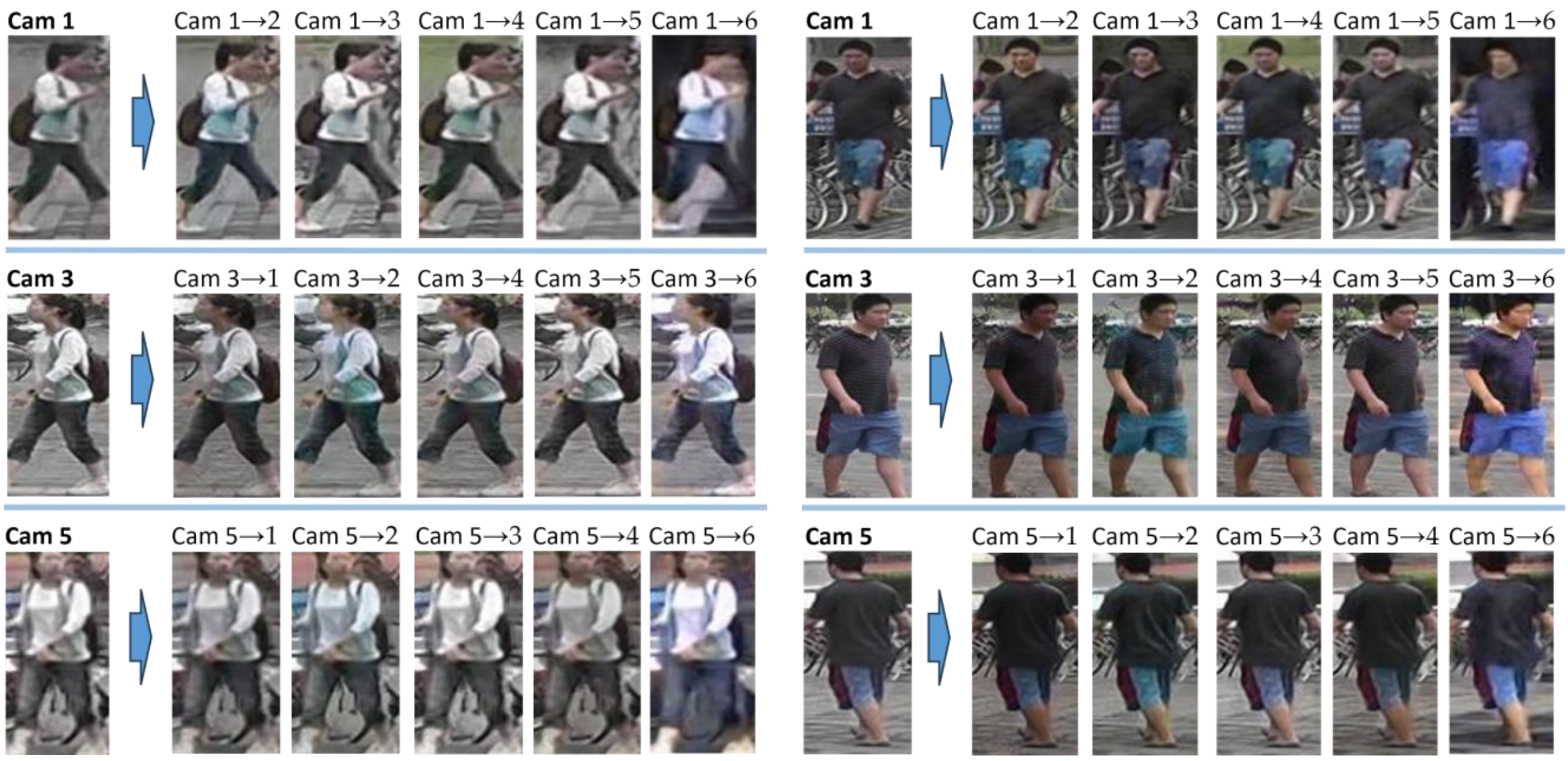

(**a**) Training data (**b**) Test data

**Figure 3.** Some style-transferred samples in Market-1501 [39]. Each image, originally taken by a specific camera, is transformed to align with the styles of the other five cameras, both within the training and test data. The real images are shown on the ***left***, while their corresponding style-transferred counterparts are shown on the ***right***.

Figure 3 shows two examples each from the training and test data in the Market-1501 dataset. The styles of these instances have been altered based on the camera view, classifying the method as a data augmentation scheme. This approach serves the dual purpose of mitigating disparities in camera styles and diminishing the impact of overfitting in Convolutional Neural Networks (CNNs). Moreover, the incorporation of camera information aids the model in learning pedestrian features with a camera-invariant property.

### *3.2. Source-domain pre-training*

#### 3.2.1. Fully supervised pre-training

As many existing the UDA approaches are based on a model pre-trained on a source dataset, our pre-training adopts a similar setup as described in [14,15,20,40]. We use ResNet101 trained on ImageNet as the backbone network (Figure 4). The last fully connected (FC) layer is removed, and two additional layers are introduced. The first layer is a batch normalization layer with 2,048 features, and the second is an FC layer with $M_S$ dimensions, where $M_S$ is the number of identities (classes) in the source dataset $S$. In our case, when training:

$$M_{S,train} = M_{S,train}^{original} + M_{S,test}^{original}, \tag{5}$$

where $M_{S,train}^{original}$ and $M_{S,test}^{original}$ are the number of identities in the original training and test sets of $S$. For each labeled image $x_{S,i}$ and its ground truth identity $y_{S,i}$ in the source domain data $\mathbb{D}_S = \{(x_{S,i}, y_{S,i})|_{i=1}^{N_S}\}$ with $N_S$ is the number of images, we train the model using the identity classification (cross-entropy) loss $\mathcal{L}_{S,ID}$ and triplet loss $\mathcal{L}_{S,triplet}$. The identity classification loss is applied to the last FC layer, treating the training process as a classification problem. Subsequently, the triplet loss is employed using the output features after batch normalization, treating the training process as a verification problem (refer to Figure 4). The loss functions are defined as follows:

$$\mathcal{L}_{S,ID} = \frac{1}{N_S}\sum_{i=1}^{N_S}\mathcal{L}_{ce}\left(C_S\left(f\left(x_{S,i}\right), y_{S,i}\right)\right), \tag{6}$$

$$\mathcal{L}_{S,triplet} = \frac{1}{N_S}\sum_{i=1}^{N_S}\max\left(0, ||f\left(x_{s,i}\right) - f\left(x_{S,i}^{+}\right)||_2 - ||f\left(x_{s,i}\right) - f\left(x_{S,i}^{-}\right)||_2 + m\right), \tag{7}$$

where $f(x_{S,i})$ is the feature of the source image $x_{S,i}$, $\mathcal{L}_{ce}$ is the cross-entropy loss, $C_S$ is a learnable source-domain classifier: $f(x_{S,i}) \rightarrow \{1,2,\ldots,M_S\}$. $||\cdot||_2$ indicates the $L^2$-norm distance, $x_{S,i}^{+}$ and $x_{S,i}^{-}$ denote the hardest positive and hardest negative feature index in each mini-batch for the sample $x_{S,i}$. The triplet distance margin is represented as $m$. With the balance parameter $\kappa$, the total loss used in source-domain pre-training is:

$$\mathcal{L}_{S,total} = \mathcal{L}_{S,ReID} = \mathcal{L}_{S,ID} + \kappa\mathcal{L}_{S,triplet}\,. \tag{8}$$

The model demonstrates good performance when trained with fully labeled data in the source domain. However, its direct application to the unlabeled target domain results in a significant drop in performance.

Prior to inputting the image into the network, we perform preprocessing by resizing the image to a specific size and applying various data augmentation techniques, including random horizontal flipping, random cropping, and edge padding. Additionally, we incorporate random color dropouts (random grayscale patch replacement) [41] to mitigate color deviation while preserving information, thereby reducing overfitting, and enhancing the model's generalization capability.

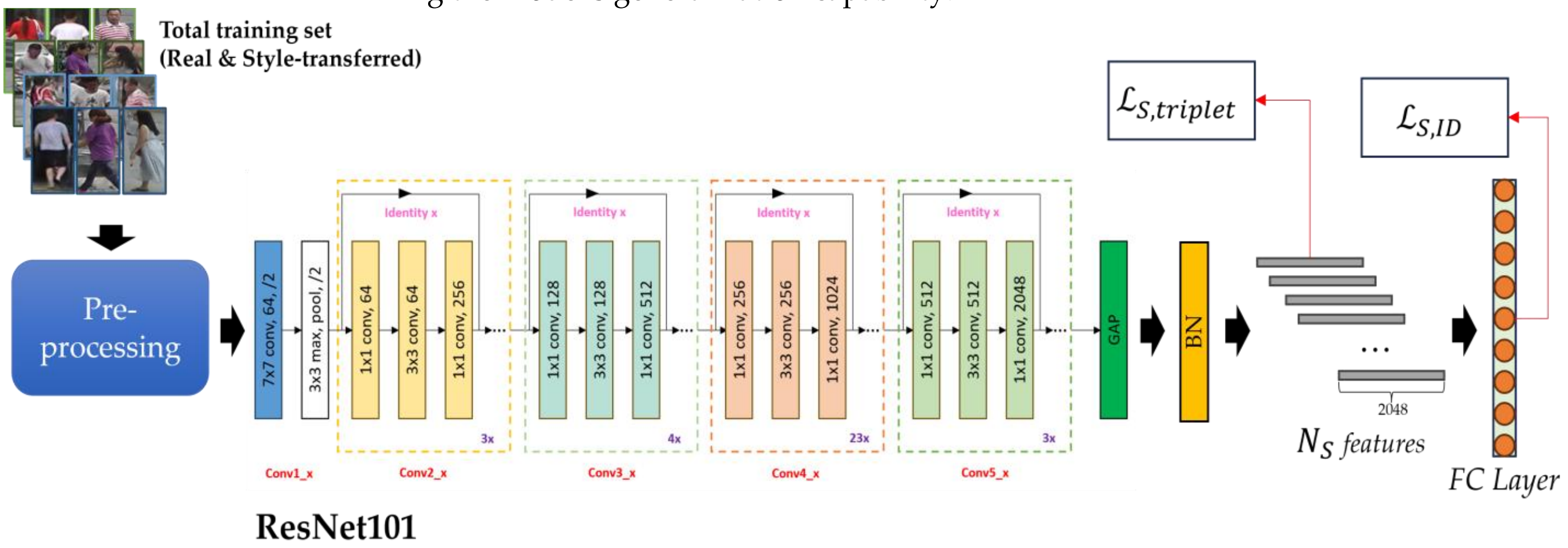

**Figure 4.** The overall training process in the fully supervised pre-training stage. ResNet101 is used as the backbone in our training process.

3.2.2. Implementation details

For the camera-aware image-to-image translation to generate synthetic data, we train 30 and 2 generative models for Market-1501 and CUHK03, respectively, using the formula $6 \times (6-1) = 30$ and $2 \times (2-1) = 2$. Throughout training, we resize all input images to 286×286 and then crop them to a size of 256× 256. We employ the Adam optimizer [42] to train the models from scratch for all experiments, setting the batch size to 8. The learning rate is initialized at 0.0002 for the Generator and 0.0001 for the Discriminator for the first 30 epochs and is linearly reduced to near zero over the remaining 20 epochs according to the lambda learning schedule policy. In the camera-aware style transfer step, we generate $C-1$ additional fake training images (5 for Market-1501 and 1 for CUHK03) while preserving their original identity, thus augmenting the training data.

As the backbone, ResNet101 is adopted, the initial learning rate is set to 0.00035 and decreased by 0.1 at the $40^{th}$ and $70^{th}$ epochs. There are a total of 120 training epochs with the number of warmup epochs is 10. We randomly sampled 32 identities and 4 images per person to form a training batch. The final batch size equals 128. In the pre-processing step, we resize each image into 256x128 pixels and pad the resized image 10 pixels using edge-padding, and then, randomly crop it into a 256 × 128 rectangular image. The augmentation methods are also applied, including random horizontal flipping of the image and random color dropouts [41] with a probability of 0.5 and 0.4, respectively. Each image is decoded into 32-bit floating point raw pixel values in $[0;1]$. Then, we normalize RGB channels by subtracting 0.485, 0.456, 0.406 and dividing by 0.229, 0.224, 0.225, respectively. The balance parameter κ is set to 1.

### *3.3. Target-domain fine-tuning*

In this phase, we use the pre-trained model to perform comprehensive optimization. We present our CORE-ReID framework (Figure 5) along with Efficient Channel Attention Block (ECAB) in the Ensemble Fusion and Bidirectional Mean Feature Normalization (BMFN) modules.

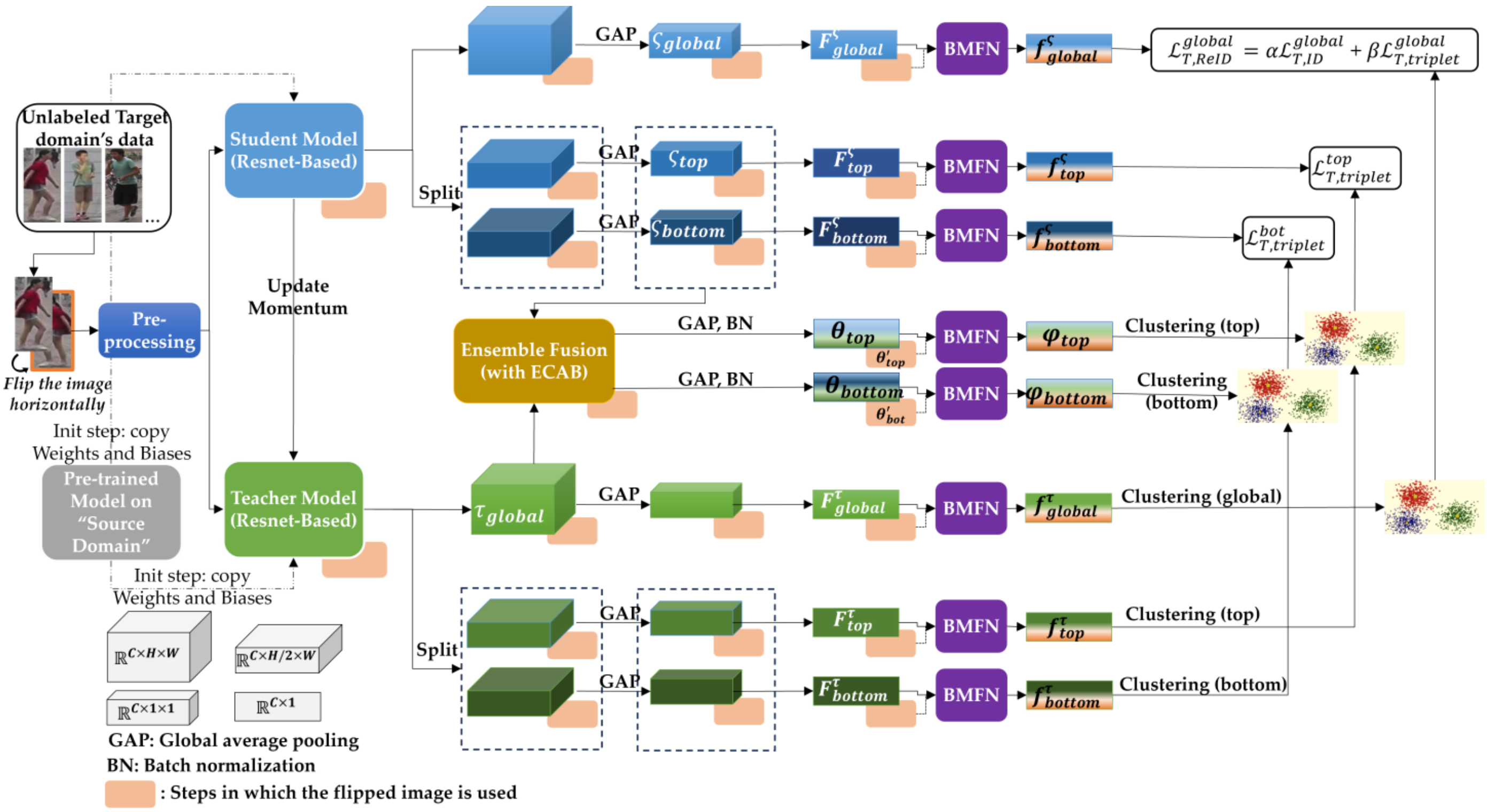

**Figure 5.** An overview of our CORE-ReID framework. We combined local features and global features using Ensemble Fusion. The ECAB in Ensemble Fusion promotes the enhancement of the features. By using BMFN, the framework can merge the feature from the original image $x_{T,i}$ and its paired flipped image $x'_{T,i}$, then produce the fusion feature $\varphi_l, l \in \{top, bottom\}$. The student network is optimized using pseudo-labels in a supervised manner, while the teacher network is updated by computing the temporal average of the student networks via the update momentum. The orange rounded rectangles indicate the steps where the features of the flipped image are used in the same way as the original image, up until the application of BMFN.

Inspired by the techniques of SSG [14] and LF² [15], our aim is to enable the model to dynamically fuse global and local (top and bottom) features, resulting in feature representations that capture both global and local information. Additionally, by constructing multiple clusters based on global and fused features, we aim to generate more consistent pseudo-labels, thus preventing ambiguous learning. To refine the noisy pseudo-labels, we adopt a pair of teacher-student networks based on the mean-teacher approach [34]. We input the same unlabeled image from the target domain to both the teacher and student networks. In the current iteration $i$, the parameters $\rho_\varsigma$ of the student network are updated using Nesterov momentum, in which, the parameters $\rho_\varsigma$ will be adjusted through backpropagation during training in the target domain Meanwhile, the parameters $\rho_\tau$ of the teacher network are computed as the temporal average of $\rho_\varsigma$, which can be expressed as:

$$\rho_{\tau,i} = \eta\rho_{\tau,i-1} + (1-\eta)\rho_\varsigma, \quad (9)$$

where $\eta$ denotes the temporal ensemble momentum, which ranges between 0 and less than 1.

3.3.1. Ensemble fusion module and overall algorithm

To derive the fusion features, we horizontally split the last global feature map of the student network into two parts (top and bottom), resulting in $\varsigma_{top}$ and $\varsigma_{bottom}$ after global average pooling. However, the last global feature map $\tau_{global}$ of the teacher network remains intact without any parting. These features, namely $\varsigma_{top}$ and $\varsigma_{bottom}$ from the student network and $\tau_{global}$ from the teacher network, are selected for adaptively learning feature fusion through the Ensemble Fusion module, which incorporates learnable parameters. In line with the approach of LF[2], we design the Ensemble Fusion module (Figure 6), wherein we initially utilize $\varsigma_{top}$ and $\varsigma_{bottom}$ along with the spatial information of the student network's local features.

The inputs $\varsigma_{top}$ and $\varsigma_{bottom}$ will be forwarded to ECAB for adaptive learning fusion. Each enhanced attention map ($\psi_{top}$ and $\psi_{bottom}$) outputted from ECAB will be merged with $\tau_{global}$ through element-wise multiplication to generate the ensemble fusion feature maps: $\tau_{global}^{top}$ and $\tau_{global}^{bottom}$. Subsequently, after applying Global Average Pooling (GAP) and batch normalization, we obtain the fusion features $\theta_{top}$ and $\theta_{bottom}$ to input into BMFN for predicting pseudo-labels using clustering algorithms in subsequent steps. The overall process in Ensemble Fusion can be expressed as:

$$\tau_{global}^{top} = \psi_{top} \otimes \tau_{global} = ECAB(\varsigma_{top}) \otimes \tau_{global}, \quad (10)$$

$$\tau_{global}^{bot} = \psi_{bottom} \otimes \tau_{global} = ECAB(\varsigma_{bottom}) \otimes \tau_{global}, \quad (11)$$

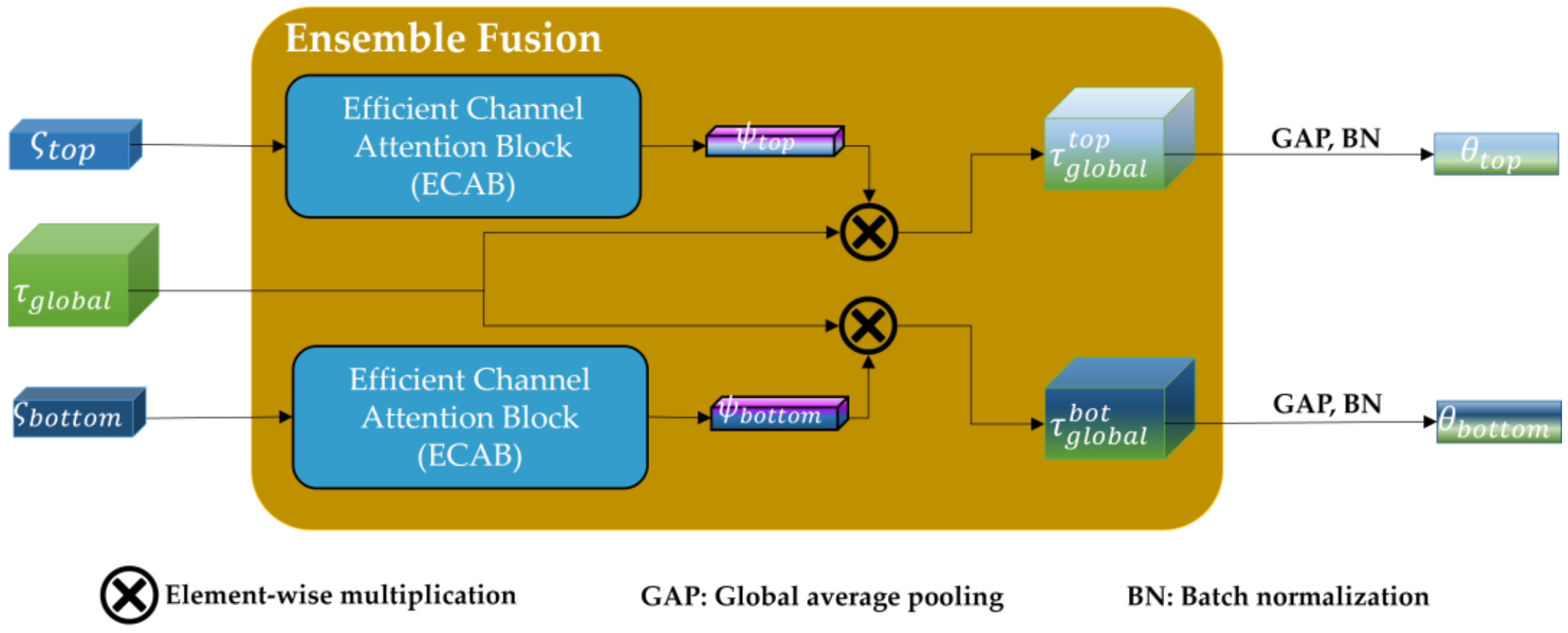

**Figure 6.** The Ensemble Fusion component. $\varsigma_{top}$ and $\varsigma_{bottom}$ features are passed through the ECAB to produce the channel attention maps by exploiting the inter-channel relationship of features which helps to enhance the features.

We apply Mini-batch K-means algorithm in clustering to predict the pseudo labels. Consequently, each $x_{T,i}$ will have three pseudo labels (global, top, and bottom respectively). We denote the target domain data as $\mathbb{D}_T = \{(x_{T,i}, \hat{y}_{T,i,j})|_{i=1}^{N_T}\}$, with $j \in \{global, top, bottom\}$ and $N_T$ is the number of images in target dataset $T$. Where $\hat{y}_{T,i,j} \in \{1,2 \ldots, M_{T,j}\}$ expresses that the pseudo label $\hat{y}_{T,i,j}$ of the target-domain image $x_{T,i}$ is from the cluster result $\hat{Y}_j = \{\hat{y}_{T,i,j} | i = 1,2, \ldots, N_T\}$ of the combined feature with its flipped image $x'_{T,i}$ outputted from BMFN: $\varphi_l, l \in \{top, bottom\}$. $M_{T,j}$ denotes the number of identities (classes) in $\hat{Y}_j$.

Before calculating the loss function, we apply the BMFN to obtain each optimized feature from networks $f_j^{\varsigma}$, $f_j^{\tau}$, $j \in \{global, top, bottom\}$ and $\varphi_l, l \in \{top, bottom\}$ from ensemble fusion. After acquiring multiple pseudo labels, we can get three new target-domain datasets for training the student network. We use the pseudo labels generated by local fusion features $\varphi_l, l \in \{top, bottom\}$ to calculate the soft-max triplet loss for corresponding local features $f_l^{\varsigma}$ from student network:

$$\mathcal{L}_{T,triplet}^{l} = \frac{1}{N_T}\sum_{i=1}^{N_T} \log\left(\frac{e^{\|f_l^{\varsigma}(x_{T,i}|\rho_{\varsigma}) - f_l^{\varsigma}(x_{T,i}^{-}|\rho_{\varsigma})\|_2}}{e^{\|f_l^{\varsigma}(x_{T,i}|\rho_{\varsigma}) - f_l^{\varsigma}(x_{T,i}^{-}|\rho_{\varsigma})\|_2} + e^{\|f_l^{\varsigma}(x_{T,i}|\rho_{\varsigma}) - f_l^{\varsigma}(x_{T,i}^{+}|\rho_{\varsigma})\|_2}}\right), \tag{12}$$

where $\rho_{\tau}$ and $\rho_{\varsigma}$ are parameters in teacher and student networks. Optimized local feature from student network is denoted as $f_l^{\varsigma}$, $l \in \{top, bottom\}$. $x_{T,i}^{+}$ and $x_{T,i}^{-}$ are the hardest positive and negative samples of the anchor target-domain image $x_{T,i}$, respectively.

Similar to supervised learning approach, we use the cluster result $\hat{Y}_{global}$ of the global clustering feature $f_{global}^{\varsigma}$ as pseudo labels to calculate the classification loss $\mathcal{L}_{T,ReID}^{global}$ and global triplet loss $\mathcal{L}_{T,triplet}^{global}$ for global feature $f_{global}^{\varsigma}$. The formulas are expressed as:

$$\mathcal{L}_{T,ID}^{global} = \frac{1}{N_T}\sum_{i=1}^{N_T} \mathcal{L}_{ce}\left(C_T\left(f_{global}^{\varsigma}(x_{T,i}), \hat{y}_{T,i,global}\right)\right), \tag{13}$$

$$\mathcal{L}_{T,triplet}^{global} = \frac{1}{N_T}\sum_{i=1}^{N_T} \max\left(0, \|f_{global}^{\varsigma}(x_{T,i}) - f_{global}^{\varsigma}(x_{T,i}^{+})\|_2 - \|f_{global}^{\varsigma}(x_{T,i}) - f_{global}^{\varsigma}(x_{T,i}^{-})\|_2 + m\right), \tag{14}$$

where $C_T$ is the fully connected layer of the student network for classification: $f_{global}^{\varsigma}(x_{T,i}) \to \{1,2,\dots,M_{T,global}\}$. $\|\cdot\|_2$ indicates the $L^2$-norm distance.

With $\alpha, \beta, \gamma$ and $\delta$ are weighting parameters, the total loss can be calculated as:

$$\begin{aligned}\mathcal{L}_{T,total} &= \mathcal{L}_{T,ReID}^{global} + \gamma\mathcal{L}_{T,triplet}^{top} + \delta\mathcal{L}_{T,triplet}^{bottom} \\ &= \alpha\mathcal{L}_{T,ID}^{global} + \beta\mathcal{L}_{T,triplet}^{global} + \gamma\mathcal{L}_{T,triplet}^{top} + \delta\mathcal{L}_{T,triplet}^{bottom}.\end{aligned} \tag{15}$$

During the inference phase, the Ensemble Fusion process is omitted, and we only use the optimized teacher network to save the computation cost. In detail, the global feature map from the teacher network is segmented into two parts, known as the top and bottom (it also serves as a student network). Following global average pooling, the resulting two local features and the global feature are concatenated. Subsequently, $L_2$ normalization and BMFN is applied to get the final optimal feature to facilitate inference.

3.3.2. ECAB

The importance of attention has been extensively explored in previous literature[43]. Attention not only guides where to focus but also enhances the representation of relevant features. Inspired by the CBAM, we introduce an ECAB, a straightforward yet impactful attention module for feed-forward convolutional neural networks. This module enhances representation power through attention mechanisms that emphasize crucial features while suppressing unnecessary ones. We generate a channel attention map by leveraging inter-channel relationships within features. Each channel of a feature map serves as a feature detector, and channel attention directs focus towards the most meaningful aspects of an input image. To efficiently compute channel attention, we compress the spatial dimension of the input feature map.

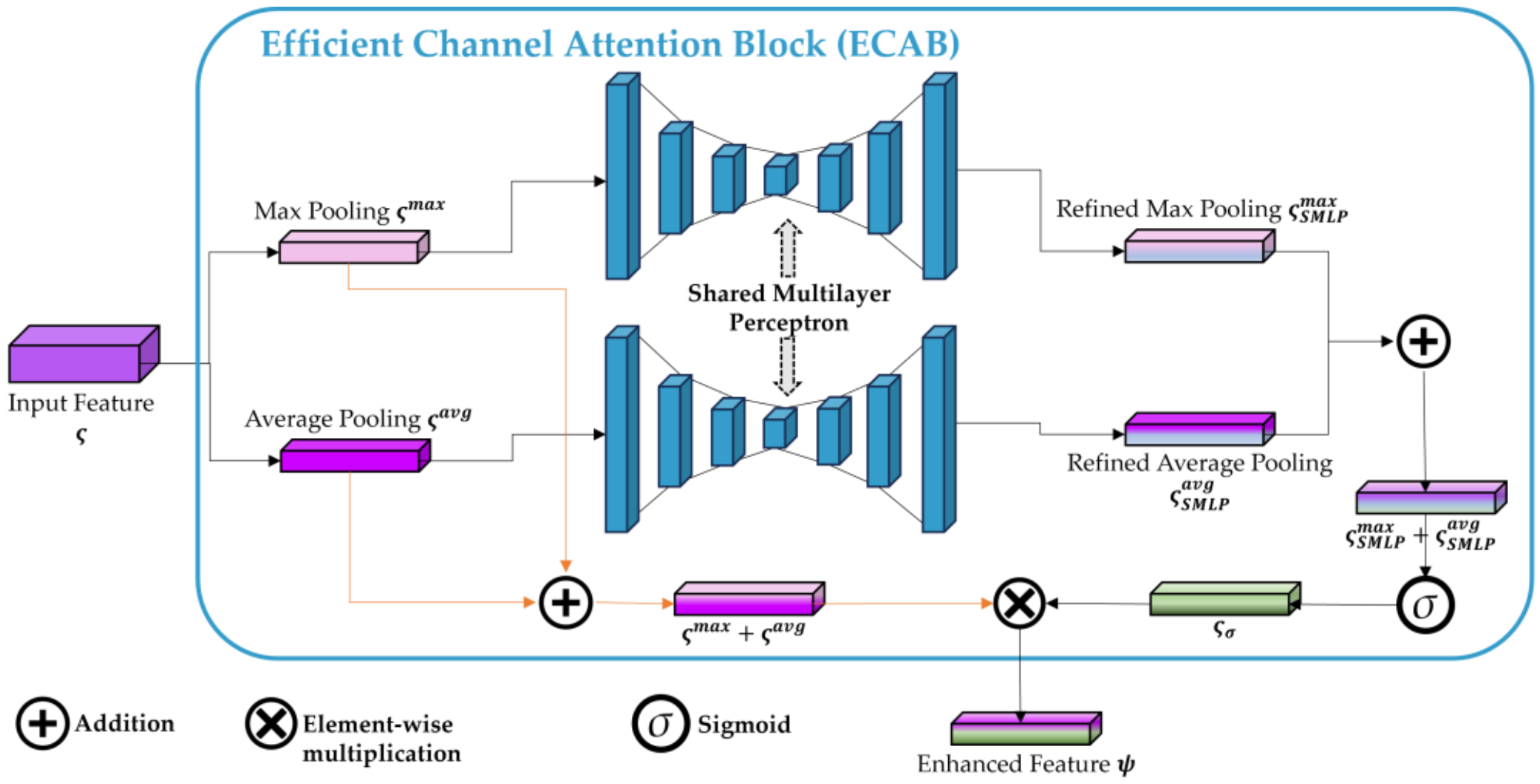

**Figure 7.** The structure of our ECAB. The Shared Multilayer Perceptron has odd $h$ hidden layers, where the first $\frac{h-1}{2}$ layers are reduced in size with the reduction rate $r$, and the last $\frac{h-1}{2}$ layers will be expanded with the same rate $r$.

While average-pooling has been commonly used to aggregate spatial information, Zhou et al. [44] suggest its effectiveness in learning object extents. Therefore, we utilize both average-pooling and max-pooling features simultaneously. Figure 7 shows the design of ECAB. Given an intermediate input feature map $\varsigma \in \mathbb{R}^{C\times W\times H}$ ,where $C, W, H$ denote the number channel, width, and height respectively. After performing max-pooling and average-pooling, then fit the outputs $\varsigma^{max}$, $\varsigma^{avg}$ into a Shared Multilayer Perceptron, we can obtain refined feature as $\varsigma_{SMLP}^{max}$ and $\varsigma_{SMLP}^{avg}$. The Shared Multilayer Perceptron has multiple hidden layers with reduction rate $r$ and the same expansion rate, activation function ReLU. The enhanced attention map $\psi \in \mathbb{R}^{C\times 1\times 1}$ is calculated as:

$$\psi = (\varsigma^{max} + \varsigma^{avg}) \otimes \varsigma_\sigma = (\varsigma^{max} + \varsigma^{avg}) \otimes \sigma(\varsigma_{SMLP}^{max} + \varsigma_{SMLP}^{avg}), \quad (16)$$

where $\varsigma_\sigma$ is the output of $\varsigma_{SMLP}^{max} + \varsigma_{SMLP}^{avg}$ after the sigmoid function.

### 3.3.3. BMFN

The use of horizontally flipped images has been studied in [45-47]. We assume that the image can be captured in the opposite direction (left and right). By using the flipped image in training, the model can focus on the id-related features and ignore the background features.

Given an image $x_{T,i}$ in target domain dataset, and its flipped image $x'_{T,i}$. After getting the feature map $F_j^m$ and its paired flipped image's feature map $F'^m_j$ , j $\in \{global, top\ bottom\}, m \in \{\varsigma, \tau\}$ . The outputs from BMFN can be calculated as:

$$f_j^m = \frac{\frac{F_j^m + F'^m_j}{2}}{||\frac{F_j^m + F'^m_j}{2}||_2}. \quad (17)$$

The optimal feature maps on the branch of Ensemble Fusion $\theta_l, l \in \{top, bottom\}$ can be intended after applying the BMFN are:

$$\varphi_l^m = \frac{\frac{\theta_l^m + \theta'^m_l}{2}}{||\frac{\theta_l^m + \theta'^m_l}{2}||_2}, \quad (18)$$

### 3.3.4. Detailed implementation

Our training regimen spans 80 epochs, with each epoch consisting of 400 iterations. Throughout training, the learning rate remains fixed at 0.00035, and we employ the Adam optimizer with a weight decay of 0.0005 to facilitate stable convergence. The utilization of the K-means algorithm aids in initializing cluster centers effectively. For temporal ensemble regularization, we set the momentum parameter (η) to 0.999. To balance the various components of our loss function, we assign the following weights: α: 1; λ: 1; γ: 0.5; δ: 0.5. In our Ensemble Fusion block, we maintain a reduction ratio and expand rate ($r$) of 4 and the number of hidden layers $h$ is set to 5. Like pre-training stage, the processing step is also applied. In addition, randomly erasing with the probability of 0.5 is conducted in fine-tuning stage.

## 4. Results

In this section, we will show the experimental results compared to SOTA methods on popular datasets for the task of UDA for Person ReID.

### *4.1. Dataset description*

We evaluate the effectiveness of our proposal by conducting evaluations on three benchmark datasets: Market-1501 [39], CUHK03 [48], and MSMT17 [49].

**Market-1501** comprises 32,668 photos featuring 1,501 individuals captured from 6 different camera views. The training set encompasses 12,936 images representing 751 identities. The testing set includes 3,368 query images and 19,732 gallery images, encompassing the remaining 750 identities.

**CUHK03** consists of 14,097 images depicting 1,467 unique identities, captured by 6 campus cameras, with each identity being recorded by 2 cameras. This dataset offers two types of annotations: manually labeled bounding boxes and those generated by an automatic detector. We employ manually annotated bounding boxes for both training and testing purposes. Additionally, we adopt a more rigorous testing protocol proposed in [40] for CUHK03. This protocol involves splitting the dataset into 767 identities (7,365 images) for training and 700 identities (5,332 images and 1400 images in gallery and query sets respectively) for testing.

**MSMT17**, a large-scale dataset, consists of 126,441 bounding boxes representing 4,101 identities captured by 12 outdoor and 3 indoor cameras (a total of 15 cameras) in three periods (morning, afternoon, and noon) throughout the day on four different days. The training set incorporates 32,621 images showcasing 1041 identities, while the testing set consists of 93,820 images featuring 3060 identities used for evaluation. The testing set is divided into 11,659 images for the query set and 82,161 images for the gallery set. Notably, MSMT17 surpasses both Market-1501 and CUHK03 in scale.

The comprehensive overview of the open datasets utilized in this document is presented in Table 1.

**Table 1.** Details of datasets used in this manuscript.

| Dataset | Cameras | Training Set (ID/Image) | Test Set (ID/Image) | |
|---|---|---|---|---|
| | | | Gallery | Query |
| Market-1501 | 6 | 751/12,936 | 750/19,732 | 750/3,368 |
| CUHK03 | 2 | 767/7,365 | 700/5,332 | 700/1,400 |
| MSMT17 | 15 | 1,401/32,621 | 3,060/82,161 | 3,060/11,659 |

**Table 2.** Experimental results of the proposed CORE-ReID framework and SOTA methods (Acc %) on Market-1501 and CUHK03 datasets. **Bold** denotes the best while Underline indicates the second-best results.

| | | Market → CUHK | | | | CUHK → Market | | | |
|---|---|---|---|---|---|---|---|---|---|
| Method | Reference | mAP | R-1 | R-5 | R-10 | mAP | R-1 | R-5 | R-10 |
| SNR [a] [50] | CVPR 2020 | 17.5 | 17.1 | - | - | 52.4 | 77.8 | - | - |
| UDAR [51] | PR 2020 | 20.9 | 20.3 | - | - | 56.6 | 77.1 | - | - |
| $QAConv_{50}$ [a] [52] | ECCV 2020 | 32.9 | 33.3 | - | - | 66.5 | 85.0 | - | - |
| $M^3L$ [a] [53] | CVPR 2021 | 35.7 | 36.5 | - | - | 62.4 | 82.7 | - | - |
| MetaBIN [a] [54] | CVPR 2021 | 43.0 | 43.1 | - | - | 67.2 | 84.5 | - | - |
| DFH-Baseline [55] | CVPR 2022 | 10.2 | 11.2 | - | - | 13.2 | 31.1 | - | - |
| DFH [a] [55] | CVPR 2022 | 27.2 | 30.5 | - | - | 31.3 | 56.5 | - | - |
| META [a] [56] | ECCV 2022 | 47.1 | 46.2 | - | - | 76.5 | 90.5 | - | - |
| ACL [a] [57] | ECCV 2022 | 49.4 | 50.1 | - | - | 76.8 | 90.6 | - | - |
| RCFA [58] | Electronics 2023 | 17.7 | 18.5 | 33.6 | 43.4 | 34.5 | 63.3 | 78.8 | 83.9 |
| CRS [59] | JSJTU 2023 | - | - | - | - | 65.3 | 82.5 | 93.0 | 95.9 |
| MTI [60] | JVCIR 2024 | 16.3 | 16.2 | - | - | - | - | - | - |
| PAOA+ [a] [61] | WACV 2024 | 50.3 | 50.9 | - | - | 77.9 | 91.4 | - | - |
| Baseline | Ours | 55.2 | 55.7 | 72.1 | 81.0 | 82.2 | 92.0 | 96.7 | 97.6 |
| CORE-ReID | Ours | **62.9** | **61.0** | **79.6** | **87.2** | **83.6** | **93.6** | **97.3** | **98.7** |

**Table 3.** Experimental results of the proposed CORE-ReID framework and SOTA methods (Acc %) from Market-1501 and CUHK03 source datasets to target domain MSMT17 dataset. **Bold** denotes the best while Underline indicates the second-best results. [a] indicates the method uses multiple source datasets, [b] denotes the implementation is based on the author's code.

| | | **Market → MSMT** | | | | **CUHK → MSMT** | | | |
|---|---|---|---|---|---|---|---|---|---|
| **Method** | **Reference** | **mAP** | **R-1** | **R-5** | **R-10** | **mAP** | **R-1** | **R-5** | **R-10** |
| NRMT [63] | ECCV 2020 | 19.8 | 43.7 | 56.5 | 62.2 | - | - | - | - |
| DG-Net++ [38] | ECCV 2020 | 22.1 | 48.4 | - | - | - | - | - | - |
| MMT [22] | ICLR 2020 | 22.9 | 52.5 | - | - | 13.5 [b] | 30.9 [b] | 44.4 [b] | 51.1 [b] |
| UDAR [51] | PR 2020 | 12.0 | 30.5 | - | - | 11.3 | 29.6 | - | - |
| Dual-Refinement [64] | arXiv 2020 | 25.1 | 53.3 | 66.1 | 71.5 | - | - | - | - |
| SNR [a] [50] | CVPR 2020 | - | - | - | - | 7.7 | 22.0 | - | - |
| $QAConv_{50}$ [a] [52] | ECCV 2020 | - | - | - | - | 17.6 | 46.6 | - | - |
| $M^3L$ [a] [53] | CVPR 2021 | - | - | - | - | 17.4 | 38.6 | - | - |
| MetaBIN [a] [54] | CVPR 2021 | - | - | - | - | 18.8 | 41.2 | - | - |
| RDSBN [65] | CVPR 2021 | 30.9 | 61.2 | 73.1 | 77.4 | - | - | - | - |
| ClonedPerson [66] | CVPR 2022 | 14.6 | 41.0 | - | - | 13.4 | 42.3 | - | - |
| META [a] [56] | ECCV 2022 | - | - | - | - | 24.4 | 52.1 | - | - |
| ACL [a] [57] | ECCV 2022 | - | - | - | - | 21.7 | 47.3 | - | - |
| CLM-Net [67] | NCA 2022 | 29.0 | 56.6 | 69.0 | 74.3 | - | - | - | - |
| CRS [59] | JSJTU 2023 | 22.9 | 43.6 | 56.3 | 62.7 | 22.2 | 42.5 | 55.7 | 62.4 |
| HDNet [68] | IJMLC 2023 | 25.9 | 53.4 | 66.4 | 72.1 | - | - | - | - |
| DDNet [69] | AI 2023 | 28.5 | 59.3 | 72.1 | 76.8 | - | - | - | - |
| CaCL [70] | ICCV 2023 | 36.5 | 66.6 | 75.3 | 80.1 | - | - | - | - |
| PAOA+ [a] [61] | WACV 2024 | - | - | - | - | 26.0 | 52.8 | - | - |
| OUDA [71] | WACV 2024 | 20.2 | 46.1 | - | - | - | - | - | - |
| M-BDA [72] | VCIR 2024 | 26.7 | 51.4 | 64.3 | 68.7 | - | - | - | - |
| UMDA [73] | VCIR 2024 | 32.7 | 62.4 | 72.7 | 78.4 | - | - | - | - |
| Baseline | Ours | 40.1 | 67.3 | 79.4 | 83.1 | 37.2 | 65.5 | 77.2 | 81.0 |
| CORE-ReID | Ours | **41.9** | **69.5** | **80.3** | **84.4** | **40.4** | **67.3** | **79.0** | **83.1** |

*4.2. Benchmark*

Our study initially focuses on comparing CORE-ReID against SOTA methods on two domain adaptation tasks: Market → CUHK and CUHK → Market (Table 2). Subsequently, we then expand our evaluation to include two additional tasks: Market → MSMT and CUHK → MSMT (Table 3). "Baseline" denotes the method that we only use the global feature, without using ECAB and BMFN, CORE-ReID denotes our framework. The evaluation metrics are mAP(%) and rank (R) at *k* accuracy (%).

The results demonstrate that our framework significantly outperforms existing SOTA methods, validating the effectiveness of our method. Specifically, by integrating the ensemble fusion component, ECAB and BMFN, our framework outperforms SOTA methods. In particular, we observe significant improvements over PAOA+, with margins of 12.6% and 6.0% mAP on both Market → CUHK and CUHK → Market tasks, respectively, despite PAOA+ incorporating additional training data.

*4.3. Ablation Study*

**Feature Maps Visualization:** To verify our method, we visualize the feature map of Grad-CAM [74] at the global feature level. Important features of each person are represented as heatmaps, as shown in Figure 8. The rainbow color describes the level of less important (blue) to most important (red) parts used for person re-identification. In the Market → CUHK and CUHK → Market scenarios (Figure 8 (a) and Figure 8 (b)), important features are observed in the target person's body. The heatmaps show identical distributions in the original and flipped images. This observation is considered to be consistent with the accuracy of our method as shown in Table 2.

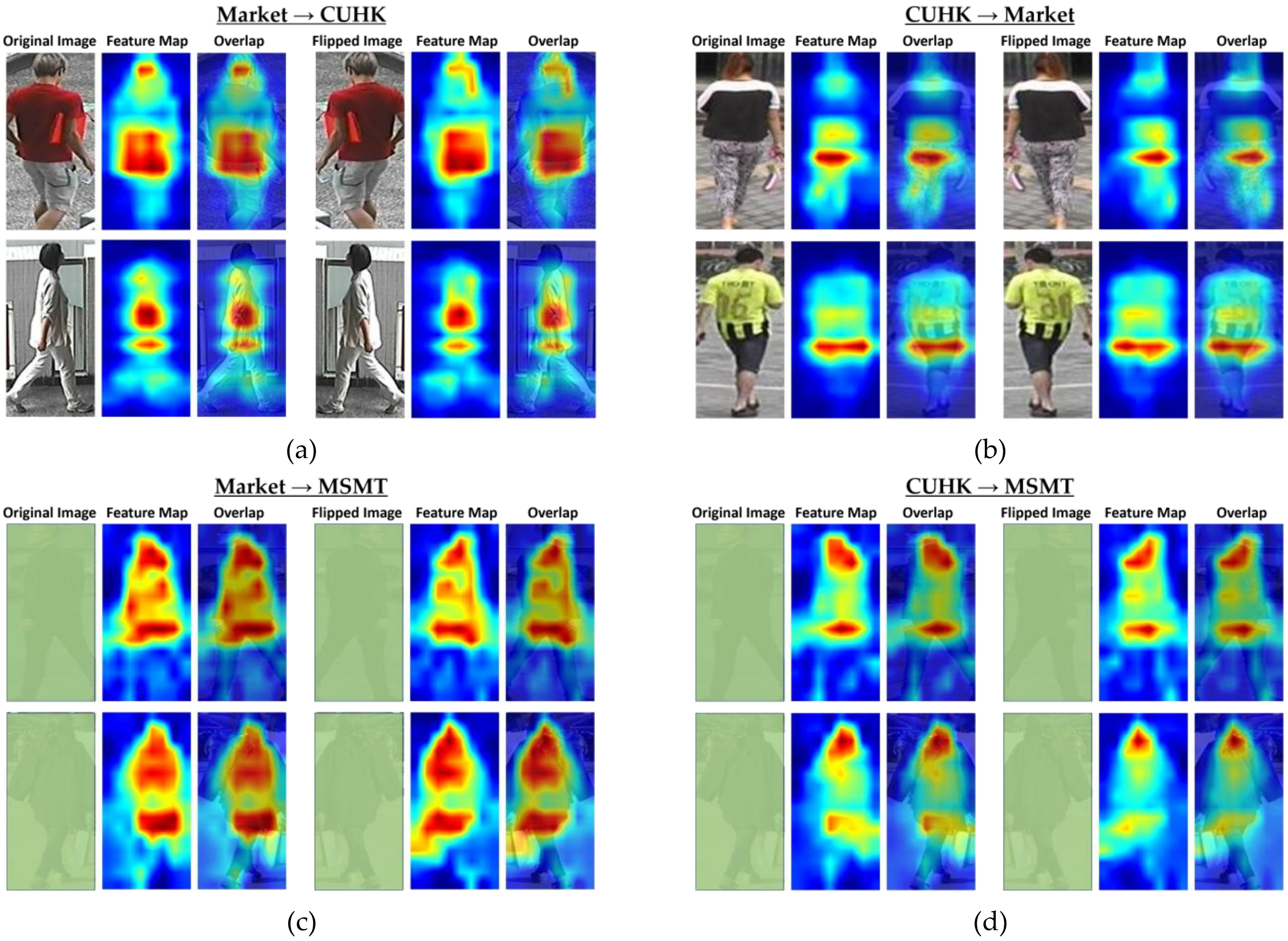

**Figure 8.** Feature maps visualization using Grad-CAM [74]. (a), (b), (c), and (d) illustrate the feature maps of those pairs on Market→CUHK, CUHK→Market, Market→MSMT, CUHK→MSMT, respectively.

On the other hand, in the Market → MSMT and CUHK→ MSMT scenarios, the Market → MSMT model shows a slightly better extraction of important features, where the heatmap is distributed in the middle and lower body regions in both the original and the flipped images. This fact could be considered as the reason for the higher accuracy achieved by the Market→ MSMT model over the CUHK→ MSMT model, as shown in Table 3.

**K-means Clustering Settings:** we use the K-means approach for clustering to make the pseudo-labels on target domain. The settings are various depending on the datasets. As shown in Table 4, our framework performs best on Market → CUHK, CUHK → Market, Market → MSMT and CUHK → MSMT with the settings of 900, 900, 2500, and 2500, respectively.

**Table 4.** Experimental results on different settings of number of pseudo identities in K-means clustering algorithm.

| | Market → CUHK | | | | CUHK → Market | | | |
|---|---|---|---|---|---|---|---|---|
| **Method** | **mAP** | **R-1** | **R-5** | **R-10** | **mAP** | **R-1** | **R-5** | **R-10** |
| Ours ($M_{T,j}$ = 500) | 52.4 | 51.4 | 70.6 | 79.1 | 77.4 | 91.0 | 96.5 | 97.6 |
| Ours ($M_{T,j}$ = 700) | 57.3 | 57.1 | 74.5 | 83.0 | 82.1 | 92.6 | **97.5** | 98.2 |
| Ours ($M_{T,j}$ = 900) | **62.9** | **61.0** | **79.6** | **87.2** | **83.6** | **93.6** | 97.3 | **98.7** |
| | **Market → MSMT** | | | | **CUHK → MSMT** | | | |
| **Method** | **mAP** | **R-1** | **R-5** | **R-10** | **mAP** | **R-1** | **R-5** | **R-10** |
| Ours ($M_{T,j}$ = 2500) | **41.9** | **69.5** | **80.3** | **84.4** | **40.4** | **67.3** | **79.0** | **83.1** |
| Ours ($M_{T,j}$ = 3000) | 39.8 | 66.8 | 78.9 | 83.0 | 37.2 | 64.7 | 76.6 | 80.9 |
| Ours ($M_{T,j}$ = 3500) | 37.6 | 65.1 | 77.3 | 81.8 | 35.0 | 63.1 | 75.4 | 79.8 |

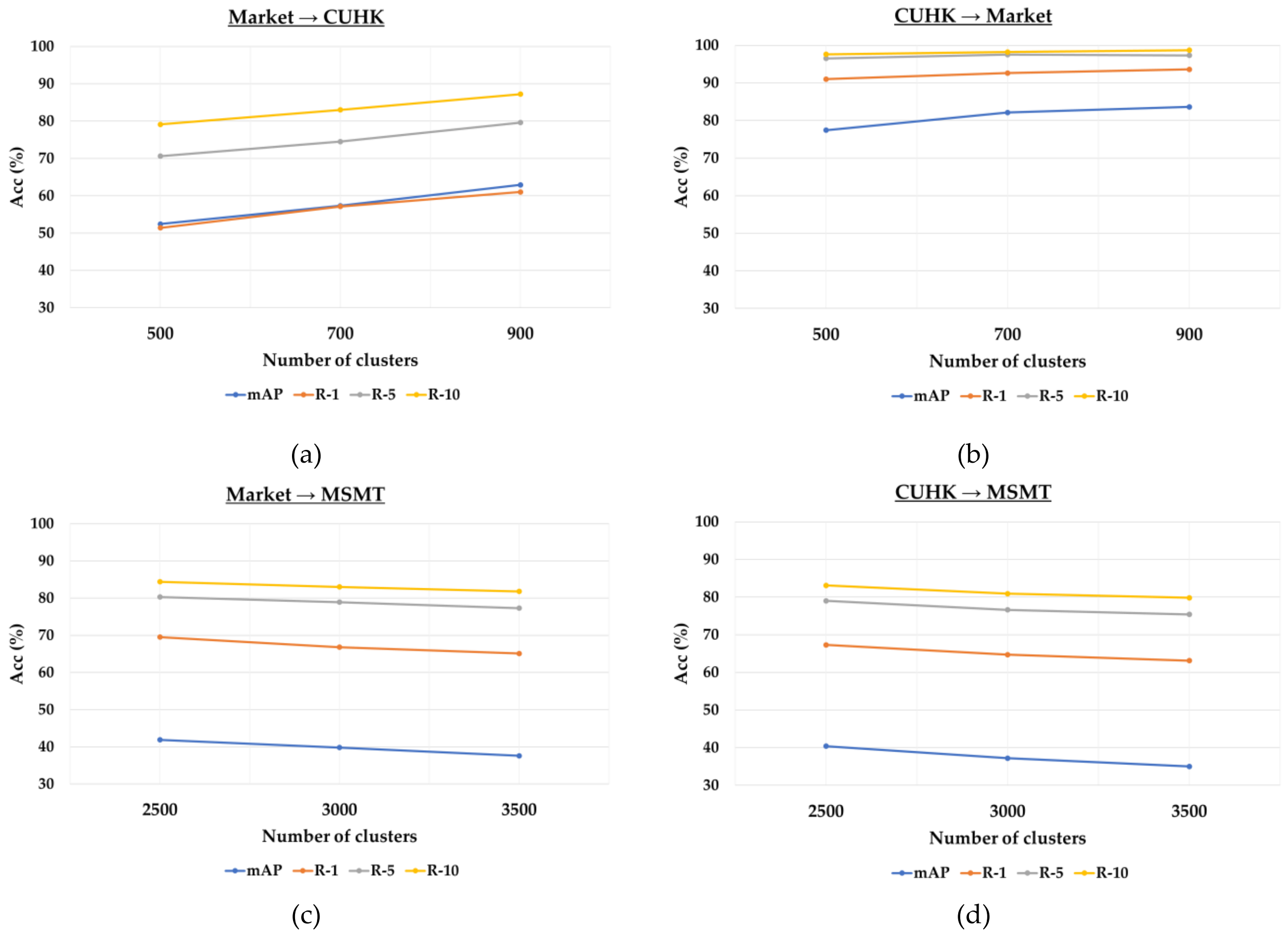

**Figure 9.** Impact of clustering parameter $M_{T,j}$. Results on (a) Market → CUHK, (b) CUHK → Market, (c) Market → MSMT, and (d) CUHK → MSMT.

Figure 9 shows that the performance of our method varies depending on the dataset pairs and the clustering parameter values ($M_{T,j}$) used.

**ECAB and BMFN Settings:** ECAM improves representation power by using attention mechanisms to highlight important features and suppress irrelevant ones. To validate the effectiveness of ECAB, we perform an experiment where it is removed from our network, as shown in Table 5.

As mentioned earlier, we use BMFN to merge features from the original image and its flipped counterpart, allowing the model to concentrate on ID-related features while disregarding background features. Table 5 demonstrates that incorporating BMFN enhances accuracy.

Figure 10 shows the results that utilizing ECAB and BMFN in our framework lead to performance improvement.

**Table 5.** Experimental results to validate the effectiveness of utilizing ECAB and BMFN in our proposed framework. The clustering parameter values ($M_{T,j}$) is carried out from the study of K-means clustering settings.

| | **Market → CUHK ($M_{T,j}$ = 900)** | | | | **CUHK → Market ($M_{T,j}$ = 900)** | | | |
|---|---|---|---|---|---|---|---|---|
| **Method** | **mAP** | **R-1** | **R-5** | **R-10** | **mAP** | **R-1** | **R-5** | **R-10** |
| Ours (without ECAB) | 56.9 | 55.8 | 72.8 | 81.6 | 83.3 | 93.4 | **97.4** | 98.4 |
| Ours (without BMFN) | 62.3 | 60.3 | 79.2 | 87.0 | 83.0 | 92.7 | 97.3 | 98.3 |
| Ours (with ECAB and BMFN) | **62.9** | **61.0** | **79.6** | **87.2** | **83.6** | **93.6** | 97.3 | **98.7** |
| | **Market → MSMT ($M_{T,j}$ = 2500)** | | | | **CUHK → MSMT ($M_{T,j}$ = 2500)** | | | |
| **Method** | **mAP** | **R-1** | **R-5** | **R-10** | **mAP** | **R-1** | **R-5** | **R-10** |
| Ours (without ECAB) | 41.2 | 68.5 | 80.1 | 83.8 | 38.0 | 65.8 | 77.5 | 81.8 |
| Ours (without BMFN) | 41.1 | 68.2 | 80.1 | 83.9 | 39.8 | 66.7 | 78.7 | 82.8 |
| Ours (with ECAB and BMFN) | **41.9** | **69.5** | **80.3** | **84.4** | **40.4** | **67.3** | **79.0** | **83.1** |

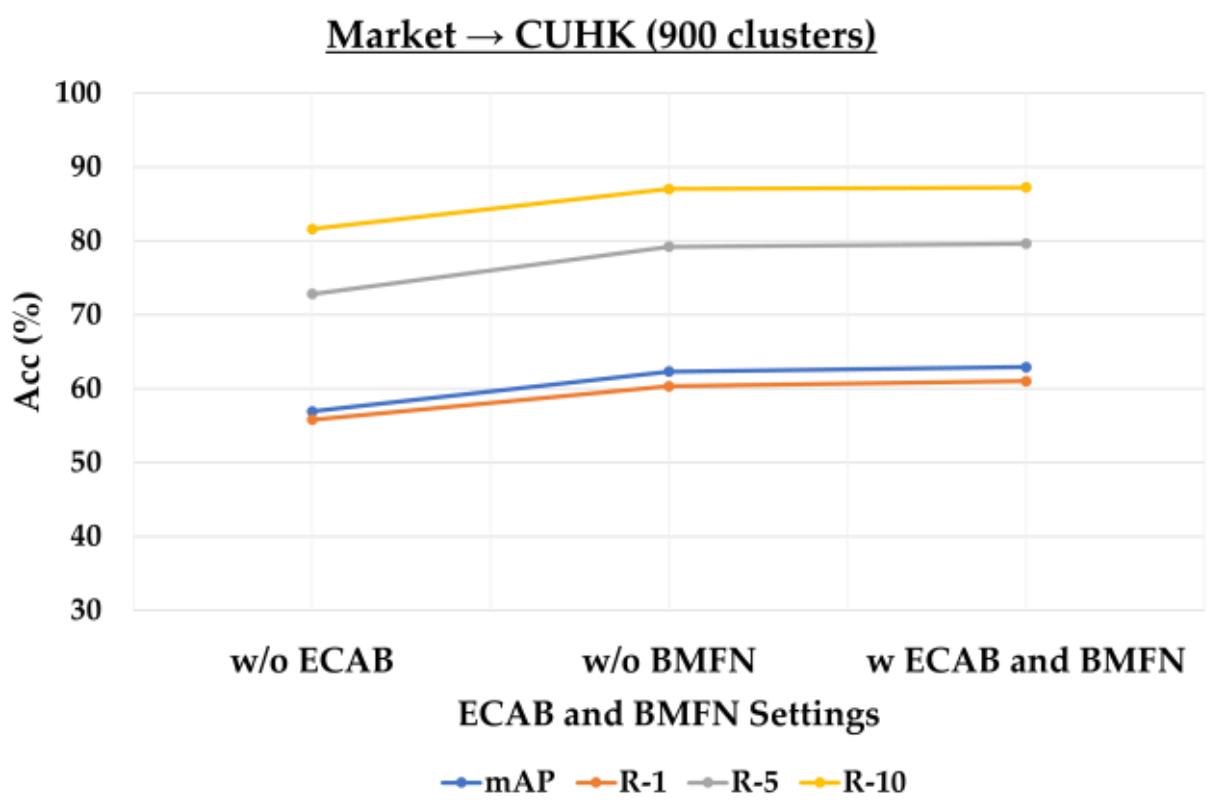

(a)

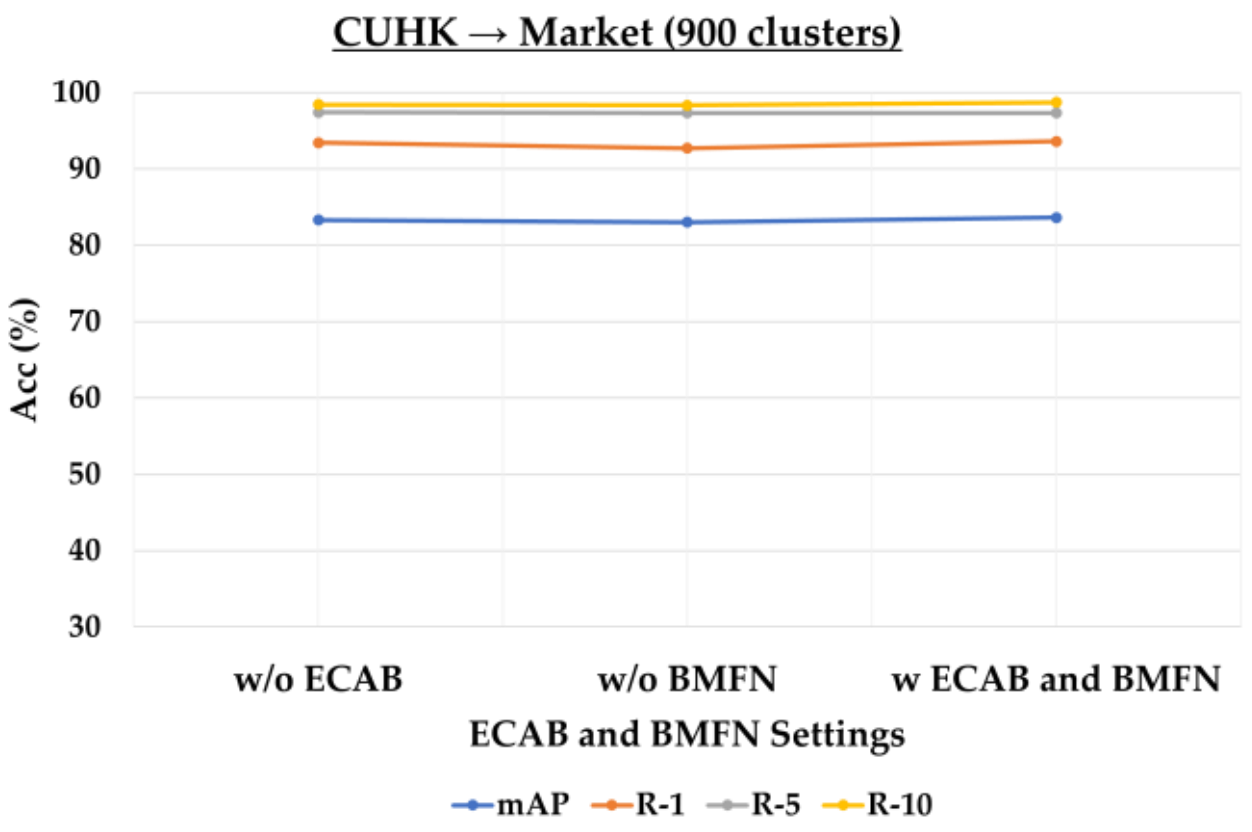

(b)

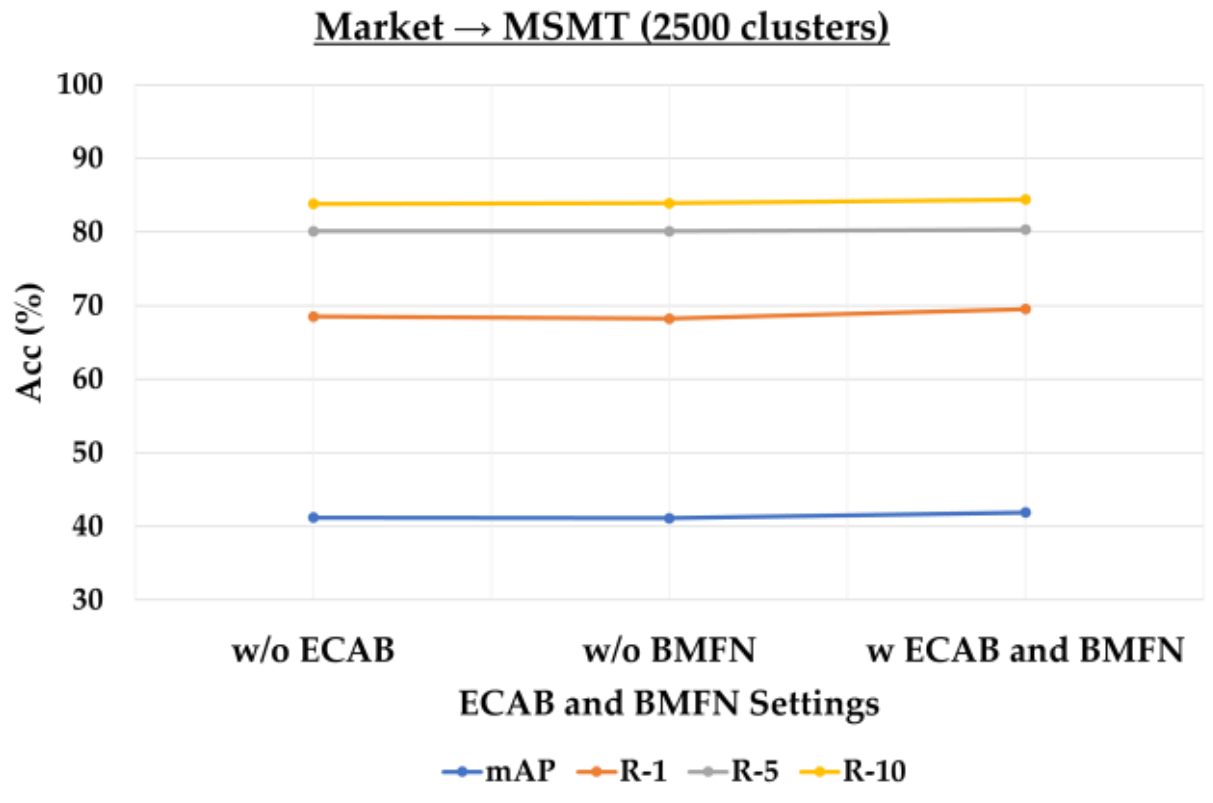

(c)

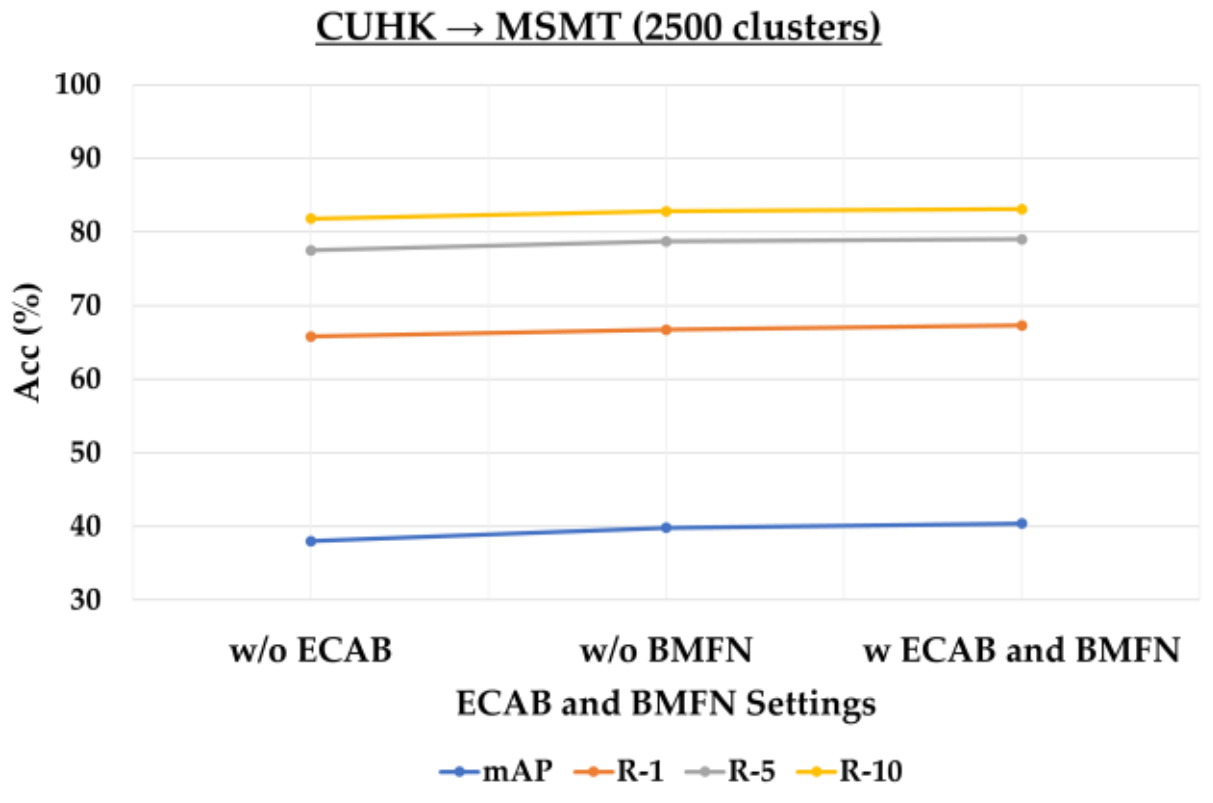

(d)

**Figure 10.** Effect of using ECAB and BMFN in our proposed methods. Results on (a) Market → CUHK, (b) CUHK → Market, (c) Market → MSMT, and (d) CUHK → MSMT.

**Backbone Configurations:** we also evaluate the performance of different backbone architectures (ResNet50, ResNet101, and ResNet152) in modeling the network for unsupervised domain adaptation in Person Re-ID. By systematically comparing the performance of these models, we aim to identify the most effective backbone architecture for our task. Through extensive experimentation and analysis, we gain valuable insights into the impact of the backbone architecture on the overall performance of the unsupervised domain adaptation framework for Person Re-ID as shown in Table 6. The ResNet101 setting gives the best performance on both Market → CUHK and CUHK → Market scenarios. All of these experiments were performed on two machines with dual Quadro RTX 8000 GPUs each.

**Table 6.** Experimental results on different settings of ResNet backbones in Market → CUHK, CUHK → Market scenarios.

| | Market → CUHK ($M_{T,j}$ = 900) | | | | CUHK → Market ($M_{T,j}$ = 900) | | | |
|---|---|---|---|---|---|---|---|---|
| **Method** | **mAP** | **R-1** | **R-5** | **R-10** | **mAP** | **R-1** | **R-5** | **R-10** |
| Ours (ResNet50) | 62.3 | 61.0 | 77.7 | 85.4 | 83.4 | 93.1 | 97.3 | 98.4 |
| Ours (ResNet101) | **62.9** | **61.0** | **79.6** | **87.2** | **83.6** | **93.6** | 97.3 | **98.7** |
| Ours (ResNet152) | 60.4 | 59.0 | 76.8 | 85.6 | 83.4 | 93.1 | **97.8** | 98.4 |

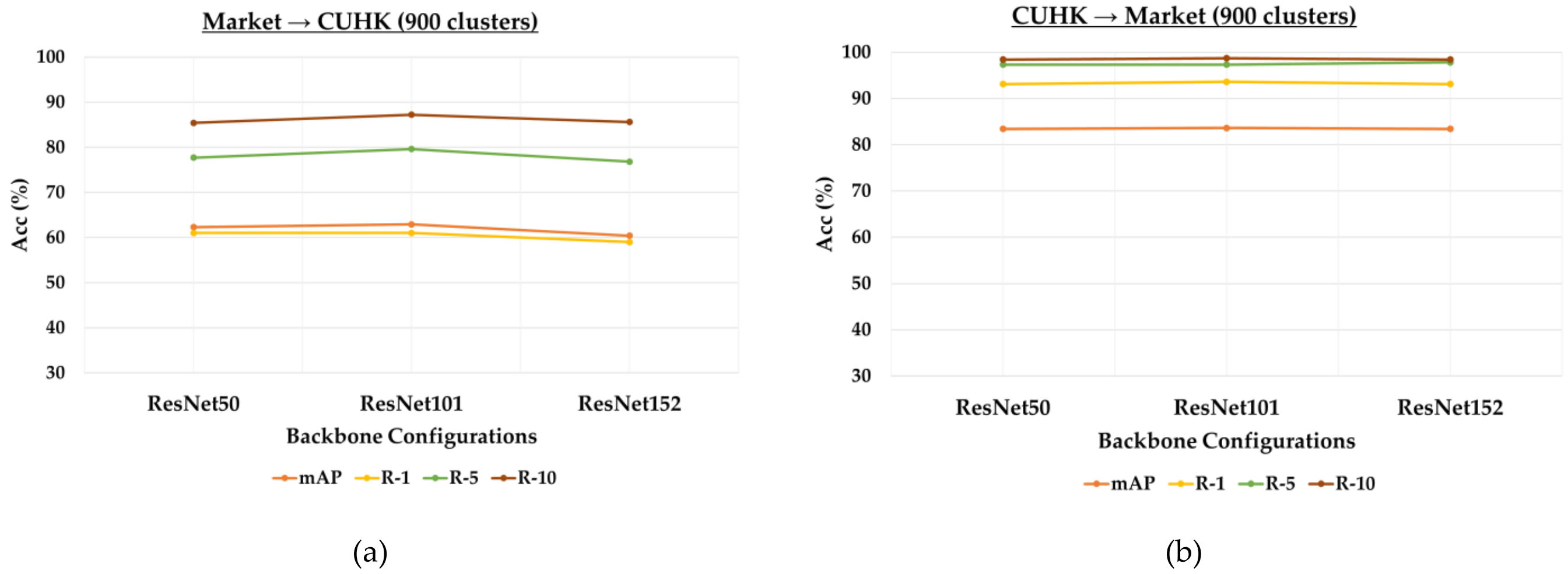

**Figure 11.** Impact of the backbone configurations. Results on (a) Market → CUHK, (b) CUHK → Market show that ResNet101 backbone gives the best overall results.

Figure 11 shows that the performance does not vary much across different backbone configurations, indicating the stability of our framework regardless of the settings used.

## 5. Conclusions

In this paper, we present a multifaceted approach to solve the problem of the UDA. Firstly, we propose a dynamic fine-tuning strategy that employs a camera-aware style transfer model to augment Re-ID data. This not only mitigates disparities in camera styles but also combats CNN overfitting on the source domain. Additionally, we introduce an Efficient Channel Attention Block (ECAB) that leverages inter-channel relationships to prioritize meaningful structures in input images, improving feature extraction. Furthermore, we establish a Comprehensive Optimization and Refinement through Ensemble fusion (CORE-ReID) framework. This framework utilizes a pair of teacher-student networks to fuse global and local features adaptively, generating diverse pseudo labels for multi-level clustering. Finally, we incorporate a Bidirectional Mean Feature Normalization (BMFN) module to enhance feature-level discriminability.

In addition to achieving SOTA performance, our method notably narrows the gap between supervised and unsupervised performance in Person Re-ID. We expect that our approach will offer valuable insights into the UDA for Person Re-ID, potentially paving the way for further advances in the field.

However, our approach has limitations. A major challenge is the dependence on the quality of the camera-aware style transfer model, which can affect the overall performance if not properly optimized. Additionally, the complexity of our CORE-ReID framework may lead to increased computational cost and training time. Future work will focus on optimizing the efficiency of the style transfer model and simplifying the CORE framework without sacrificing performance. We also plan to explore more advanced techniques for noise reduction in pseudo-labels to further enhance the robustness of our model.

**Author Contributions:** Writing—original draft, T.Q.N., O.D.A.P.; Conceptualization, T.Q.N. and O.D.A.P.; methodology, T.Q.N.; software, T.Q.N., O.D.A.P. and K.H.; validation, T.Q.N., O.D.A.P. and K.H.; formal analysis, T.Q.N.; investigation, T.Q.N. and O.D.A.P.; resources, T.Q.N.; data

curation, T.Q.N. and K.H.; writing—review and editing, T.Q.N., O.D.A.P. and K.H.; visualization, T.Q.N. and O.D.A.P.; supervision, O.D.A.P.; project administration, T.Q.N.; funding acquisition, T.Q.N. All authors have read and agreed to the published version of the manuscript.

**Funding:** This research received no external funding.

**Institutional Review Board Statement:** Not applicable.

**Informed Consent Statement**: Not applicable.

**Data Availability Statement:** Data is contained within the article.

**Acknowledgments:** The authors express their gratitude to CyberCore Co., Ltd for providing the training environments and the anonymous reviewers who contributed to raising the paper's standard. The authors thank Raymond Swannack, a PhD candidate student at the Graduate School of Information and Computer Science Iwate Prefectural University for proofreading our manuscript.

**Conflicts of Interest:** The authors declare that they have no conflicts of interest.

## Abbreviations

The following abbreviations are used in this manuscript:

| | |
|---|---|
| ECAB | Efficient Channel Attention Block |
| BMFN | Bidirectional Mean Feature Normalization |
| CBAM | Convolutional Block Attention Module |
| CNN | Convolutional Neural Network |
| CORE-ReID | Comprehensive Optimization and Refinement through Ensemble fusion in Domain Adaptation for person re-identification |
| HHL | Hetero-Homogeneous Learning |
| MMFA | Multi-task Mid-level Feature Alignment |
| MMT | Mutual Mean-Teaching |
| ReID | Person Re-identification |
| SOTA | State-Of-The-Art |
| SSG | Self-Similarity Grouping |
| UDA | Unsupervised Domain Adaptation |
| UNRN | Uncertainty-Guided Noise-Resilient Network |

## References

1. Zheng, L.; Yang, Y.; Hauptmann, A.G. Person Re-identification: Past, Present and Future. *arXiv* **2016**, doi:arXiv:1610.02984.
2. Ye, M.; Shen, J.; Lin, G.; Xiang, T.; Shao, L.; Hoi, S.C.H. Deep Learning for Person Re-Identification: A Survey and Outlook. *IEEE* **2022**, *44*, 2872-2893, doi:10.1109/TPAMI.2021.3054775.
3. Liu, J.; Zha, Z.J.; Xie, H.; Xiong, Z.; Zhang, Y. CA3Net: Contextual-Attentional Attribute-Appearance Network for Person Re-Identification. *Proceedings of the 26th ACM international conference on Multimedia* **2018**, 737–745, doi:10.1145/3240508.3240585.
4. Zhang, L.; Xiang, T.; Gong, S. Learning a Discriminative Null Space for Person Re-identification. *Proceedings of the IEEE Conference on Computer Vision and Pattern Recognition* **2016**, 1239–1248.
5. Somers, V.; Vleeschouwer, C.D.; Alahi, A. Body Part-Based Representation Learning for Occluded Person Re-Identification. *WACV* **2023**, 1613-1623, doi:10.1109/WACV56688.2023.00166.
6. Ghosh, A.; Shanmugalingam, K.; Lin, W.Y. Relation Preserving Triplet Mining for Stabilising the Triplet Loss in Re-identification Systems. *arXiv* **2023**, doi:arXiv:2110.07933.
7. Wang, X.; Zheng, W.S.; Li, X.; Zhang, J. Cross-Scenario Transfer Person Reidentification. *IEEE Transactions on Circuits and Systems for Video Technology* **2016**, *26*, 1447-1460, doi:10.1109/TCSVT.2015.2450331.
8. Chang, X.; Ma, Z.; Wei, X.; Hong, X.; Gong, Y. Transductive semi-supervised metric learning for person re-identification. *Pattern Recognition* **2020**, *108*, doi:10.1016/j.patcog.2020.107569.
9. Yu, H.X.; Wu, A.; Zheng, W.-S. Cross-view Asymmetric Metric Learning for Unsupervised Person Re-identification. In Proceedings of the ICCV, 2017.
10. Yu, H.X.; Wu, A.; Zheng, W.S. Cross-view Asymmetric Metric Learning for Unsupervised Person Re-identification. *ICCV* **2017**.
11. Liu, M.-Y.; Breuel, T.; Kautz, J. Unsupervised Image-to-Image Translation Networks. *NIPS* **2017**.
12. Fan, H.; Zheng, L.; Yang, Y. Unsupervised Person Re-identification: Clustering and Fine-tuning. *arXiv* **2017**, doi:arXiv:1705.10444.
13. Zhai, Y.; Lu, S.; Ye, Q.; Shan, X.; Chen, J.; Ji, R.; Tian, Y. Adcluster: Augmented discriminative clustering for domain adaptive person re-identification. *CVPR* **2020**.
14. Fu, Y.; Wei, Y.; Wang, G.; Zhou, Y.; Shi, H.; Huang, T.S. Self-similarity Grouping: A Simple Unsupervised Cross Domain Adaptation Approach for Person Re-identification. *ICCV* **2019**, doi:10.1109/ICCV.2019.00621.
15. Ding, J.; Zhou, X. Learning Feature Fusion for Unsupervised Domain Adaptive Person Re-identification. *ICPR* **2022**, doi:10.1109/ICPR56361.2022.9956264.
16. Woo, S.; Park, J.; Lee, J.-Y.; Kweon, I.S. CBAM: Convolutional Block Attention Module. *ECCV* **2018**.
17. Chen, H.; Lagadec, B.; Bremond, F. Enhancing diversity in teacher-student networks via asymmetric branches for unsupervised person re-identification. *WACV* **2021**.
18. Deng, W.; Zheng, L.; Ye, Q.; Kang, G.; Yang, Y.; Jiao, J. Image-Image Domain Adaptation with Preserved Self-Similarity and Domain-Dissimilarity for Person Re-identification. *CVPR* **2018**, 994–1003.
19. Li, Y.-J.; Lin, C.-S.; Lin, Y.-B.; Wang, Y.-C.F. Cross-Dataset Person Re-Identification via Unsupervised Pose Disentanglement and Adaptation. *ICCV* **2019**, 7919–7929.
20. Zhong, Z.; Zheng, L.; Li, S.; Yang, Y. Generalizing A Person Retrieval Model Hetero-and Homogeneously. *ECCV* **2018**, 172-188.
21. Lin, Y.; Dong, X.; Zheng, L.; Yan, Y.; Yang, Y. A Bottom-Up Clustering Approach to Unsupervised Person Re-Identification. *AAAI* **2019**, *33*, 8738-8745, doi:https://doi.org/10.1609/aaai.v33i01.33018738.
22. Ge, Y.; Chen, D.; Li, H. Mutual Mean-Teaching: Pseudo Label Refinery for Unsupervised Domain Adaptation on Person Re-identification. *ICLR* **2020**.

23. Zheng, K.; Lan, C.; Zeng, W.; Zhang, Z.; Zha, Z.-J. Exploiting Sample Uncertainty for Domain Adaptive Person Re-Identification. *AAAI* **2021**.
24. Wang, J.; Zhu, X.; Gong, S.; Li, W. Transferable joint attributeidentity deep learning for unsupervised person re-identification. *CVPR* **2018**.
25. Lin, S.; Li, H.; Li, C.-T.; Kot, A.C. Multi-task Mid-level Feature Alignment Network for Unsupervised Cross-Dataset Person Re-Identification. *BMVC* **2018**.
26. Qi, L.; Wang, L.; Huo, J.; Zhou, L.; Shi, Y.; Gao, Y. A Novel Unsupervised Camera-aware Domain Adaptation Framework for Person Re-identification. *ICCV* **2019**.
27. Wu, A.; Zheng, W.-S.; Lai, J.-H. Unsupervised Person Re-Identification by Camera-Aware Similarity Consistency Learning. *ICCV* **2019**.
28. Gou, J.; Yu, B.; Maybank, S.J.; Tao, D. Knowledge Distillation: A Survey. *IJCV* **2021**.
29. Zhang, L.; Song, J.; Gao, A.; Chen, J.; Bao, C.; Ma, K. Be Your Own Teacher: Improve the Performance of Convolutional Neural Networks via Self Distillation. *ICCV* **2019**.
30. Beyer, L.; Zhai, X.; Royer, A.; Markeeva, L.; Anil, R.; Kolesnikov, A. Knowledge distillation: A good teacher is patient and consistent. *CVPR* **2022**.
31. Laine, S.; Aila, T. Temporal ensembling for semi-supervised learning. *ICLR* **2017**.
32. Zhang, W.; Ouyang, W.; Li, W.; Xu, D. Collaborative and adversarial network for unsupervised domain adaptation. *CVPR* **2018**.
33. Zhai, Y.; Ye, Q.; Lu, S.; Jia, M.; Ji, R.; Tian, Y. Multiple Expert Brainstorming for Domain Adaptive Person Re-identification. *ECCV* **2020**.
34. Tarvainen, A.; Valpola, H. Mean teachers are better role models: Weight-averaged consistency targets improve semi-supervised deep learning results. *Advances in Neural Information Processing Systems* **2017**, *30*.
35. Zhu, J.Y.; Park, T.; Isola, P.; Efros, A.A. Unpaired image-to-image translation using cycle-consistent adversarial networks. *ICCV* **2017**.
36. Zhong, Z.; Zheng, L.; Zheng, Z.; Li, S.; Yang, Y. Camera Style Adaptation for Person Re-identification. *CVPR* **2018**.
37. Taigman, Y.; Polyak, A.; Wolf, L. Unsupervised Cross-Domain Image Generation. *ICLR* **2017**.
38. Zou, Y.; Yang, X.; Yu, Z.; Bhagavatula, V.; Kautz, J. Joint Disentangling and Adaptation for Cross-Domain Person Re-Identification. *ECCV* **2020**.
39. Zheng, L.; Shen, L.; Tian, L.; Wang, S.; Wang, J.; Tian, Q. Scalable Person Re-identification: A Benchmark. *ICCV* **2015**, 1116–1124, doi:10.1109/ICCV.2015.133.
40. Zhong, Z.; Zheng, L.; Cao, D.; Li, S. Re-ranking Person Re-identification with k-Reciprocal Encoding. *CVPR* **2017**, doi:10.1109/CVPR.2017.389.
41. Chen, Y.G.L.H.L. Eliminate Deviation with Deviation for Data Augmentation and a General Multi-modal Data Learning Method. *arXiv* **2021**.
42. Kingma, D.P.; Ba, J.L. Adam: A method for stochastic optimization. *ICLR* **2015**.
43. Itti, L.; Koch, C.; Niebur, E. A Model of Saliency-based Visual Attention for Rapid Scene Analysis. *Pattern Analysis and Machine Intelligence* **1998**, *20*, 1254-1259, doi:10.1109/34.730558.
44. Zhou, B.; Khosla, A.; Lapedriza, A.; Oliva, A.; Torralba, A. Learning Deep Features for Discriminative Localization. *CVPR* **2016**, 2921–2929.
45. Ni, X.; Rahtu, E. FlipReID: Closing the Gap between Training and Inference in Person Re-Identification. *CVPR* **2021**.
46. He, L.; Liao, X.; Liu, W.; Liu, X.; Cheng, P.; Mei, T. FastReID: A Pytorch Toolbox for General Instance Re-identification. *CVPR* **2006**.
47. Wang, G.; Yuan, Y.; Chen, X.; Li, J.; Zhou, X. Learning discriminative features with multiple granularities for person re-identification. *Proceedings of the 26th ACM international conference on Multimedia* **2018**, 274-282.

48. Li, W.; Zhao, R.; Xiao, T.; Wang, X. DeepReID: Deep Filter Pairing Neural Network for Person Re-Identification. *CVPR* **2014**.
49. Wei, L.; Zhang, S.; Gao, W.; Tian, Q. Person transfer gan to bridge domain gap for person reidentification. *CVPR* **2018**, 79-88, doi:10.1109/CVPR.2018.00016.
50. Jin, X.; Lan, C.; Zeng, W.; Chen, Z.; Zang, L. Style normalization and restitution for generalizable person re-identification. *CVPR* **2020**.
51. Song, L.; Wang, C.; Zhang, L.; Du, B.; Zhang, Q.; Huang, C.; Wang, X. Unsupervised domain adaptive re-identification: Theory and practice. *Pattern Recognition* **2020**, *102*, doi:https://doi.org/10.1016/j.patcog.2019.107173.
52. Liao, S.; Shao, L. Interpretable and generalizable person re-identification with query-adaptive convolution and temporal lifting. *ECCV* **2020**.
53. Zhao, Y.; Zhong, Z.; Yang, F.; Luo, Z.; Lin, Y.; Li, S.; Sebe, N. Learning to generalize unseen domains via memory-based multi-source meta-learning for person re-identification. *CVPR* **2021**.
54. Choi, S.; Kim, T.; Jeong, M.; Park, H.; Kim, C. Meta batch-instance normalization for generalizable person re-identification. *CVPR* **2021**.
55. Yang, F.; Zhong, Z.; Luo, Z.; Li, S.; Sebe, N. Federated and Generalized Person Re-identification through Domain and Feature Hallucinating. *CVPR* **2022**.
56. Xu, B.; Liang, J.; He, L.; Sun, Z. Mimic Embedding via Adaptive Aggregation: Learning Generalizable Person Re-identification. *ECCV 2022* **2022**, 372-388.
57. Zhang, P.; Dou, H.; Yu, Y.; Li, X. Adaptive Cross-Domain Learning for Generalizable Person Re-Identification. *ECCV* **2022**.
58. Li, H.; Wang, Y.; Zhu, L.; Wang, W.; Yin, K.; Li, Y.; Yin, G. Weakly Supervised Cross-Domain Person Re-Identification Algorithm Based on Small Sample Learning. *Electronics* **2023**, doi:https://doi.org/10.3390/electronics12194186.
59. Yanmei, M.; Huafeng, L.; Yafei, Z. Unsupervised Domain Adaptation for Cross-Regional Scenes Person Re-identification. *Journal of Shanghai Jiao Tong University* **2023**.
60. Sun, J.; Li, Y.; CHen, L.; CHen, H.; Peng, W. Multiple integration model for single-source domain generalizable person re-identification *Journal of Visual Communication and Image Representation* **2024**, *98*, doi:https://doi.org/10.1016/j.jvcir.2023.104037.
61. Li, Q.; Gong, S. Mitigate Domain Shift by Primary-Auxiliary Objectives Association for Generalizing Person ReID. *WACV* **2024**, 393-402, doi:10.1109/WACV57701.2024.00046.
62. Wang, D.; Zhang, S. Unsupervised Person Re-Identification via Multi-Label Classification. *CVPR* **2020**, 10981-10990.
63. Zhao, F.; Liao, S.; Xie, G.-S.; Zhao, J.; Zhang, K.; Shao, L. Unsupervised domain adaptation with noise resistible mutual-training for person re-identification. *ECCV* **2020**, 526-544.
64. Dai, Y.; Liu, J.; Bai, Y.; Tong, Z.; Duan, L.-Y. Dual-Refinement: Joint Label and Feature Refinement for Unsupervised Domain Adaptive Person Re-Identification. *arXiv* **2020**.
65. Bai, Z.; Wang, Z.; Wang, J.; Hu, D.; Ding, E. Unsupervised Multi-Source Domain Adaptation for Person Re-Identification. *CVPR* **2021**, 12914-12923.
66. Wang, Y.; Liang, X.; Liao, S. Cloning Outfits from Real-World Images to 3D Characters for Generalizable Person Re-Identification. *CVPR* **2022**.
67. Tay, C.-P.; Yap, K.-H. Collaborative learning mutual network for domain adaptation in person re-identification. *Neural Computing and Applications* **2022**, *34*, 12211–12222.
68. Hua, Z.; Jun, K.; Min, J.; Tianshan, L. Heterogeneous dual network with feature consistency for domain adaptation person re-identification. *International Journal of Machine Learning and Cybernetics* **2023**, *14*, 1951–1965.
69. Xiao, Y.; Qunqun, W.; Xiaozhou, C.; Kaili, S.; Yanjing, S. Discrepant mutual learning fusion network for unsupervised domain adaptation on person re-identification. *Applied Intelligence* **2023**, *53*, 2951–2966.
70. Lee, G.; Lee, S.; Kim, D.; Shin, Y.; Yoon, Y.; Ham, B. Camera-Driven Representation Learning for Unsupervised Domain Adaptive Person Re-identification. *ICCV* **2023**, 1453-11462.

71. Rami, H.; Giraldo, J.H.; Winckler, N.; Lathuilière, S. Source-Guided Similarity Preservation for Online Person Re-Identification. *WACV* **2024**, 1711-1720.
72. Zhang, B.; Wu, D.; Lu, X.; Li, Y.; Gu, Y.; Li, J.; Wang, J. A domain generalized person re-identification algorithm based on meta-bond domain alignment. *Journal of Visual Communication and Image Representation* **2024**, *98*, 2913–2933, doi:https://doi.org/10.1016/j.jvcir.2024.104054.
73. Tian, Q.; Cheng, Y.; He, S.; Sun, J. Unsupervised multi-source domain adaptation for person re-identification via feature fusion and pseudo-label refinement. *Computers and Electrical Engineering* **2024**, *113*, doi:https://doi.org/10.1016/j.compeleceng.2023.109029.
74. Selvaraju, R.R.; Cogswell, M.; Das, A.; Vedantam, R.; Parikh, D.; Batra, D. Grad-CAM: Visual Explanations from Deep Networks via Gradient-based Localization. *IJCV* **2019**.